\pdfoutput=1
\documentclass[letterpaper]{article} 
\usepackage[]{aaai25}

\usepackage{verbatim}
\usepackage{times}  
\usepackage{helvet}  
\usepackage{courier}  
\usepackage[hyphens]{url}  
\usepackage{graphicx} 
\usepackage{natbib}  
\usepackage{caption} 
\usepackage{amssymb}
\usepackage{url}
\usepackage{booktabs}
\usepackage{graphicx}
\usepackage{float}
\usepackage{multirow}
\usepackage{xcolor}
\usepackage{colortbl}
\usepackage{xspace}
\usepackage{tablefootnote}
\usepackage{tabularx}
\usepackage{pdflscape}
\usepackage{subcaption}
\usepackage{amsmath}
\usepackage{parskip}

\usepackage{ltablex}

\usepackage{algorithm}
\usepackage{algorithmic}

\definecolor{lightblue}{rgb}{0.93,0.95,1.0}
\definecolor{lightgreen}{rgb}{0.93,1.0,0.95}
\definecolor{darkgreen}{HTML}{00A64F}
\definecolor{red}{HTML}{FF0000}

\newcommand{\green}[1]{{\textbf{\color{darkgreen} #1}}}
\newcommand{\red}[1]{{\textbf{\color{red} #1}}}

\title{Beyond Retrieval: Generating Narratives in Conversational Recommender Systems}

\author{Krishna Sayana\thanks{Correspondence to: ksayana@google.com},
Raghavendra Vasudeva,
Yuri Vasilevski,
Kun Su,
Liam Hebert\thanks{Work done while an intern at Google}, \\
James Pine,
Hubert Pham,
Ambarish Jash,
Sukhdeep Sodhi
}

\affiliations{
    Google Research, Mountain View, California, USA
}

\nocopyright
\begin{document}
\maketitle

\begin{abstract}
The recent advances in Large Language Model’s generation and reasoning capabilities present an opportunity to develop truly conversational recommendation systems. However, effectively integrating recommender system knowledge into LLMs for natural language generation which is tailored towards  recommendation tasks remains a challenge. This paper addresses this challenge by making two key contributions.

First, we introduce a new dataset (REGEN) for natural language generation tasks in conversational recommendations. REGEN (\textbf{R}eviews \textbf{E}nhanced with \textbf{GE}nerative \textbf{N}arratives) extends the Amazon Product Reviews dataset with rich user narratives, including personalized explanations of product preferences, product endorsements for recommended items, and summaries of user purchase history. REGEN is made publicly available to facilitate further research. Furthermore, we establish benchmarks using well-known generative metrics, and perform an automated evaluation of the new dataset using a rater LLM. 
Second, the paper introduces a fusion architecture (CF model with an LLM) which serves as a baseline for REGEN. And to the best of our knowledge, represents the first attempt to analyze the capabilities of LLMs in understanding recommender signals and generating rich narratives. We demonstrate that LLMs can effectively learn from simple fusion architectures utilizing interaction-based CF embeddings, and this can be further enhanced using the metadata and personalization data associated with items. Our experiments show that combining CF and content embeddings leads to improvements of 4-12\% in key language metrics compared to using either type of embedding individually. We also provide an analysis to interpret how CF and content embeddings contribute to this new generative task.
\end{abstract}

\section{Introduction}
The Web is evolving towards richer and more interactive experiences, and conversational recommender systems are at the forefront of this evolution. These systems leverage natural language generation to provide personalized recommendations in the form of engaging narratives, enhancing user interaction and satisfaction. This paper introduces REGEN, a dataset specifically created to advance research in this area and improve the Web's ability to provide personalized and interactive recommendations.

Large Language Models (LLMs) have ushered in a new era of possibilities in natural language processing, enabling machines to generate human-quality text, translate languages, write different kinds of creative content, and answer user questions in an informative way.  This in turn has led to a surge of interest in applying LLMs to various domains, including the long-standing challenge of building conversational recommender systems \cite{lin2021m6, zhang2022generating, jannach2020survey}. Conversational recommender systems aim to improve upon traditional recommender systems by engaging users in dialogue to better understand and cater to their preferences. While traditional systems simply provide a list of items, conversational systems could explain recommendations, ask clarifying questions, and create narratives connecting items to user preferences. 

However, realizing this vision requires overcoming a fundamental challenge: ensuring that the generated language is relevant to both the recommendation itself and the user's preferences. This necessitates effectively integrating a latent representation of the user and the recommender's internal state with the LLM. Traditional approaches, such as collaborative filtering, excel at leveraging user-item interaction data to predict preferences and provide recommendations \cite{koren2009matrix}. Yet, they often fall short in generating the rich, informative, and engaging language that characterizes human interactions.  This gap between the structured world of recommender systems and the nuanced world of human language presents a significant hurdle.

Bridging this gap requires a careful consideration of how to represent and integrate user interaction and item information into the LLM architecture. This poses several research and design questions that need to be addressed. Should we rely on external retrieval mechanisms to fetch relevant information, or should we directly encode user preferences and item representations into the LLM?  How can we ensure that the generated language is not only informative but also aligned with the user's individual preferences and the underlying recommendation? Is it better to separately scale the recommender and language models, enabling efficient handling of large item spaces while benefiting from advancements in language modeling or should they be co-trained? How well do the fusion architectures similar to multimodal language models work on conversational recommendation tasks? 

To help answer the questions articulated above, we believe that the research community will benefit from datasets specifically designed for conversational tasks. These data\-sets should include narratives that accurately reflect user preferences and behavior within the context of recommendations. However, many existing datasets either concentrate on next item recommendations, structured outputs or short summaries that lack the rich and varied conversational elements necessary for effectively training and evaluating the capabilities of these systems.

To accelerate research in this direction, we present REGEN (\textbf{R}eviews \textbf{E}nhanced with \textbf{GE}nerative \textbf{N}arratives). REGEN augments the Amazon Reviews dataset \cite{ni-etal} by incorporating natural language outputs relevant to conversational recommendations. This augmented dataset now includes diverse examples such as purchase reasons, explanations, product endorsements, user summaries \& concise user profiles and available for public use~\cite{regendata}.
Furthermore, we use an auto-rater LLM (\cite{longpre2024foundational}) to assess the generated outputs across multiple attributes, with a particular focus on enhancing grounding and factuality based on the user's historical interactions. In our work, we use a Gemini Pro model with few shot prompting instead of finetuning as in \cite{longpre2024foundational}. Our aim is to deliver a comprehensive framework and dataset that can serve as a foundation for further research.

Further, we introduce a task and framework for conversational recommender systems: generating rich natural language outputs from user-item interaction signals that are consistent with user preferences. This is made possible by the REGEN dataset presented in this paper, and requires models to go beyond traditional recommendation tasks and produce engaging and informative narratives tailored to individual users.

To establish a strong baseline for this task, we propose a fusion architecture that seamlessly integrates collaborative filtering signals and content embeddings as input to an LLM. This approach allows the LLM to leverage both user-item interaction data and rich item content representations to generate outputs that are not only informative but also aligned with user preferences. Our experiments demonstrate the effectiveness of this fusion strategy in producing human-like conversational recommendations, showcasing its potential for creating more engaging and personalized user experiences.

Our key contributions can be summarized as follows:

\begin{enumerate}
    \item We generate a new dataset with a diverse set of narratives suitable for language recommendation tasks. The dataset is available for public use. It has been thoroughly evaluated for factuality, grounding, and accurately capturing user preferences and context . 
    \item We propose an efficient, and scalable architecture fusing embeddings from collaborative filtering and content representations as inputs to an LLM. 
    \item We use this to benchmark the dataset and establish key language metrics using the state-of-the-art LLMs. We conduct ablation studies to demonstrate the importance of the combined CF and semantic signals on the quality of the narrative outputs.
    \item We show that the model learns to construct rich in-context and aggregate narratives from history instead of simply memorizing using several examples.
    \item We analyze the soft token embeddings to determine how the model combines content and collaborative filtering signals when generating rich narratives.
\end{enumerate} 

\section{Related Work}

\subsection{Conversational Datasets}
Several datasets have been used to explore the application of LLMs in recommender systems. ReDial \cite{li2017redial} is an annotated dataset of dialogues where users recommend movies to each other, consisting of over 10,000 conversations. MIND \cite{wu2020mind} is a large-scale dataset for news recommendation and personalization, containing news articles and user interactions for conversational news recommendation scenarios. E-ConvRec \cite{chen2022econvrec} provides a large-scale conversational recommendation dataset for e-commerce customer service, based on pre-sales dialogues. TG-ReDial \cite{zhang2021topic} extends ReDial by incorporating topic information to guide conversations and recommendations. DuRecDial \cite{liu2022durecdial} is another conversational recommendation dataset based on the DuReader question-answering dataset, where users ask questions about products and receive recommendations. Some focus on short-form recommendations, while others lack the diverse conversational elements needed to evaluate recommender LLMs for generating extended, natural language responses.

The "Justifying Recommendations" dataset by Ni et al. \cite{ni-etal} and by Chen et al. \cite{chen2024unlockingwhybuyingintroducing} also focus on generating explanations from reviews. Motivated by these works, we further extend to more open ended narratives, prompting the LLM with the entire user history, and covering use cases beyond purchase explanations. Our work targets natural language outputs that reflect the rapidly evolving capabilities of LLMs, with longer, more nuanced, and potentially semi-structured or unstructured responses, both in context and in aggregate that target conversational recommenders.

\subsection{Recommender LMs}
While various approaches have been proposed to integrate LLMs with recommender systems, they can be broadly classified into three main categories:

\textbf{I. Retrieval-Augmented Generation (RAG)}:
RAG enhances traditional recommender systems by employing a retrieval model to select relevant items from a catalog based on user context. A generation model, often an LLM, then produces the final recommendations. However, RAG's effectiveness hinges on the retrieval model's accuracy, which may not be fully personalized. Moreover, RAG can introduce computational overhead during inference, potentially impacting real-time performance. Evaluating RAG systems also presents challenges in balancing compute costs and recommendation quality. Furthermore, RAG systems cannot be trained end-to-end using a gradient based approach.
Notable works exploring RAG approaches for LLMs include Lewis et al. \cite{lewis2020retrieval}, Izacard and Grave \cite{izacard2021leveraging}, Guu et al. \cite{guu2020realm}, Borgeaud et al. \cite{borgeaud2021improving}, Nakano et al. \cite{nakano2021webgpt}, Thoppilan et al. \cite{thoppilan2022lamda}, and Izacard et al. \cite{izacard2022atlas}.

\textbf{II. Fully Language-Based Techniques}
Representing items and user interactions solely through language presents several challenges for recommender systems.  While this approach allows for the use of powerful LLMs, it relies heavily on the richness of available text descriptions.  Moreover, encoding user signals at the token level can lead to a loss of  information compared to item-level encoding.  Scalability is also a concern, as processing long interaction histories becomes computationally expensive.  Finally, incorporating additional metadata, which is often crucial for effective recommendations, poses a significant hurdle.
Examples of this approach include M6-Rec by Lin et al. \cite{lin2021m6}, 
CALRec~\cite{calrec24} by Li et al. and the work of Zhang et al. \cite{zhang2022generating} .The RecSys 2022 workshop \cite{recsys2022language} further highlights this trend.

\textbf{III. Embedding Input-Based Language Models}
This approach focuses on utilizing item representations as direct embeddings inputs to an LLM~\cite{hebert2024persomapersonalizedsoftprompt,tennenholtz2024embeddingalignedlanguagemodels, ning2024userllmefficientllmcontextualization, doddapaneni2024userembeddingmodelpersonalized, ilm}. 
However, the generation is restricted to categorical variables such as genre or product categories \cite{hebert2024persomapersonalizedsoftprompt} and some of the studies generate up to a single sentence, like a concise summary \cite{ning2024userllmefficientllmcontextualization}, which do not explore the generative capabilities of these systems. This work expands on those efforts by introducing conversational tasks that require generating more complex and nuanced narratives.  Our analysis demonstrates that these tasks provide valuable insights into how LLMs learn and utilize recommender signals. 
Note that the embedding based approaches are further motivated by the soft-prompt based parameter efficient tuning of language models proposed in ~\cite{li2021prefixtuningoptimizingcontinuousprompts, lester2021powerscaleparameterefficientprompt}. 

We use the architecture motivated by III, which allows us to easily incorporate the upstream embeddings from a purely collaborative model, as well as embeddings from a semantic understanding model seamlessly. Further, this results in far fewer tokens as input to the LM, as opposed to using text representations for history, thereby allowing faster serving in practice. 
While this independent encoding strategy is advantageous for our current investigation and improves reproducibility, future work should explore whether jointly training the encoders with the LM or fusing these embeddings on a recommendations task could yield further improvements, especially when dealing with high-dimensional embedding spaces with large item counts common in production datasets. Notably, several recent approaches have sought to enhance recommendation quality by incorporating item and/or semantic context, including hybrid models like HybridBERT4Rec \cite{sun2020bert4rec}, FLARE \cite{hebert2024flarefusinglanguagemodels} and CALRec \cite{calrec24}, self-supervised methods like S$^3$Rec \cite{wu2021s3rec}, intent-based models like CaFe \cite{qian2021cafe}, and models leveraging cross-attention \cite{liu2022carca}, pre-training \cite{qi2020user}, semantic item IDs and T5-style training \cite{yuan2022tiger,raffel2020exploring} or aligning CF and content representations \cite{kddLLMCF, kddBridgingItemsLanguage}.

\begin{figure*}[!htb]
    \centering
    \begin{subfigure}{0.4\textwidth}
        \includegraphics[width=\linewidth]{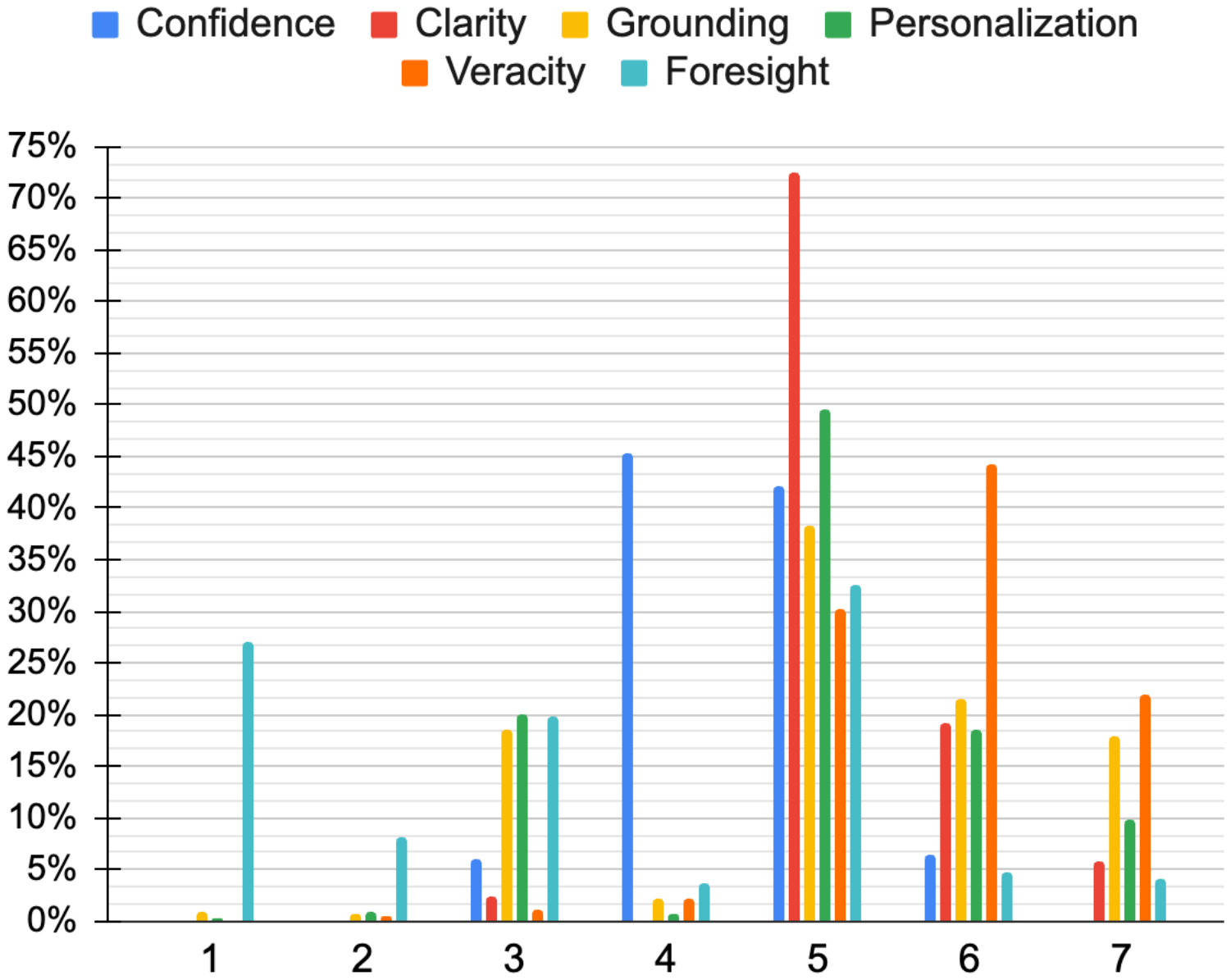}
        \caption{Office Products}
    \end{subfigure}
    \hspace{5mm}
    \begin{subfigure}{0.4\textwidth}
        \includegraphics[width=\linewidth]{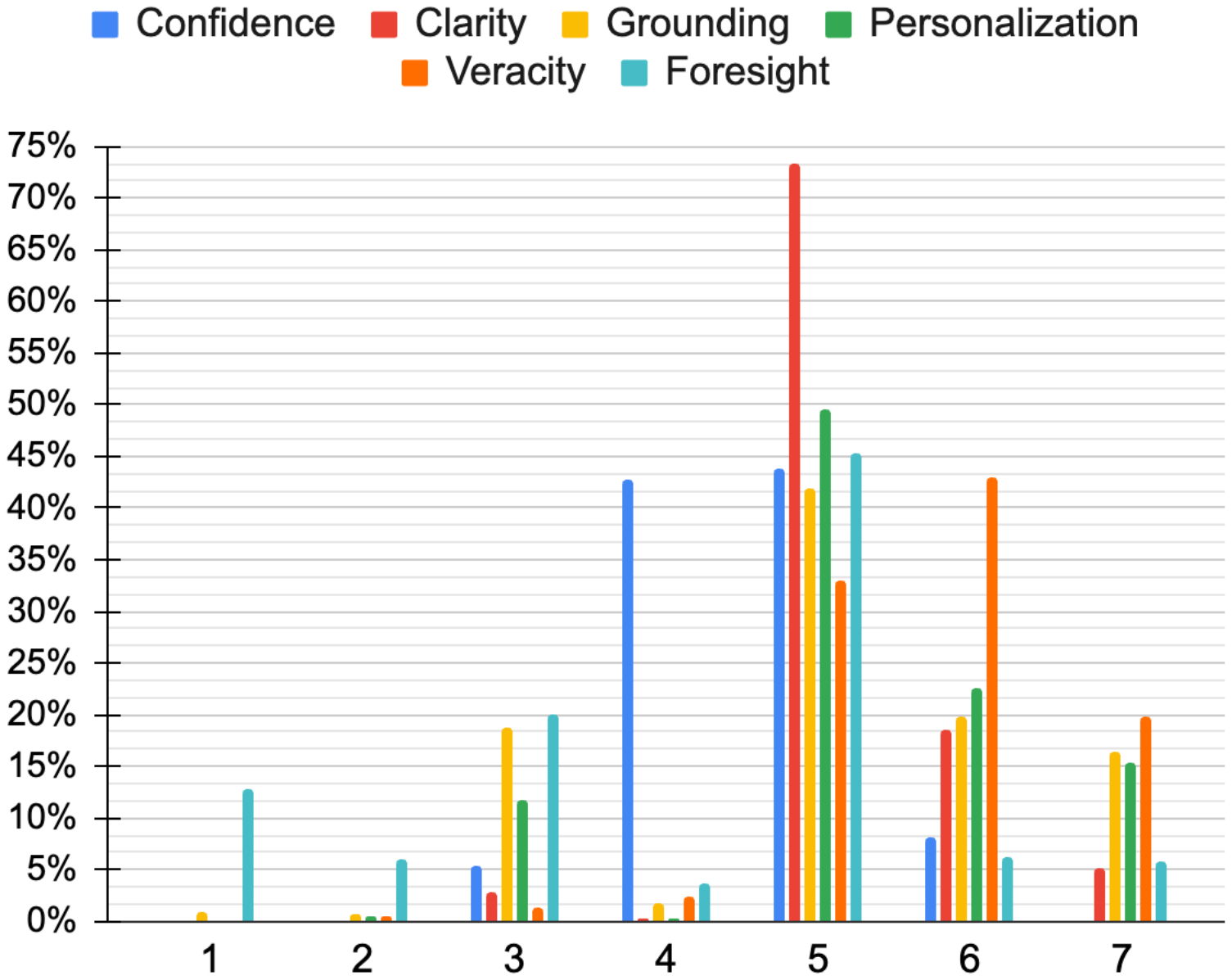}
        \caption{Clothing, Jewelry \& Accessories}
    \end{subfigure}
    \caption{Purchase reason by rated attributes.}
    \label{fig:pre_eval}
\end{figure*}

\begin{figure*}[!htb]
    \centering
    \begin{subfigure}{0.4\textwidth}
        \includegraphics[width=\linewidth]{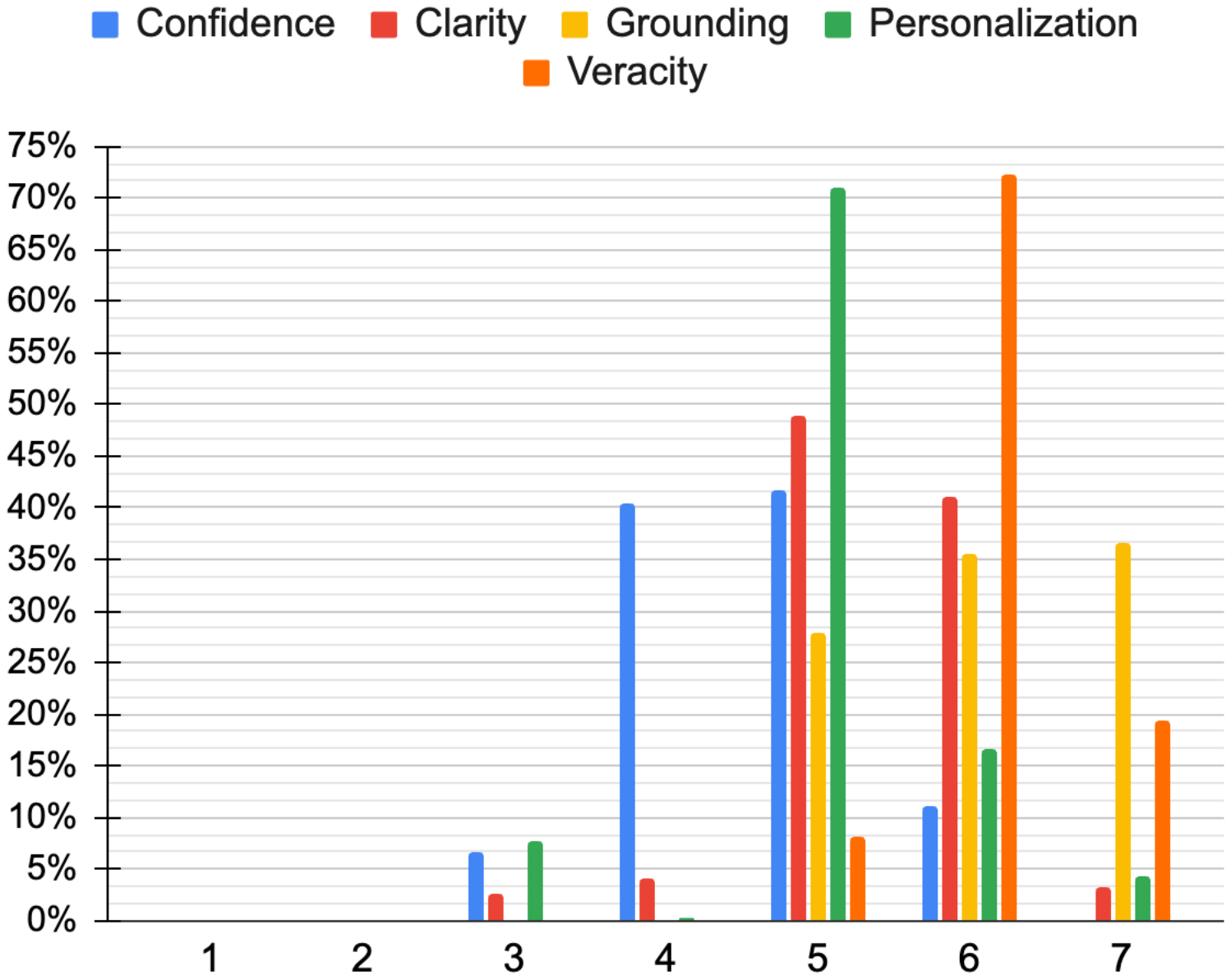}
        \caption{Office Products}
    \end{subfigure}
    \hspace{5mm}
    \begin{subfigure}{0.4\textwidth}
        \includegraphics[width=\linewidth]{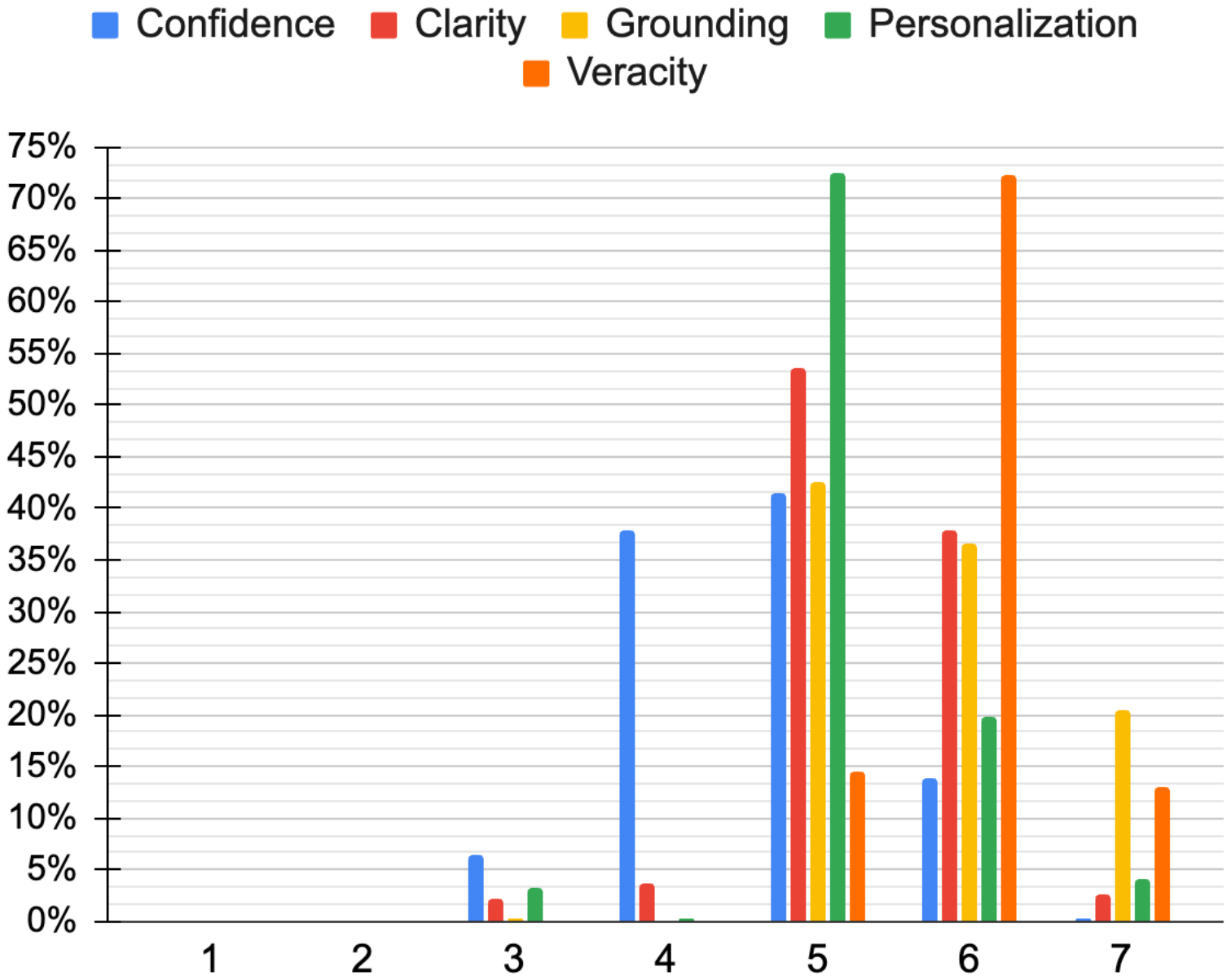}
        \caption{Clothing, Jewelry \& Accessories}
    \end{subfigure}
    \caption{User summary by rated attributes.}
    \label{fig:us_eval}
\end{figure*}

\begin{figure*}[!htb]
    \centering
    \begin{subfigure}{0.4\textwidth}
        \includegraphics[width=\linewidth]{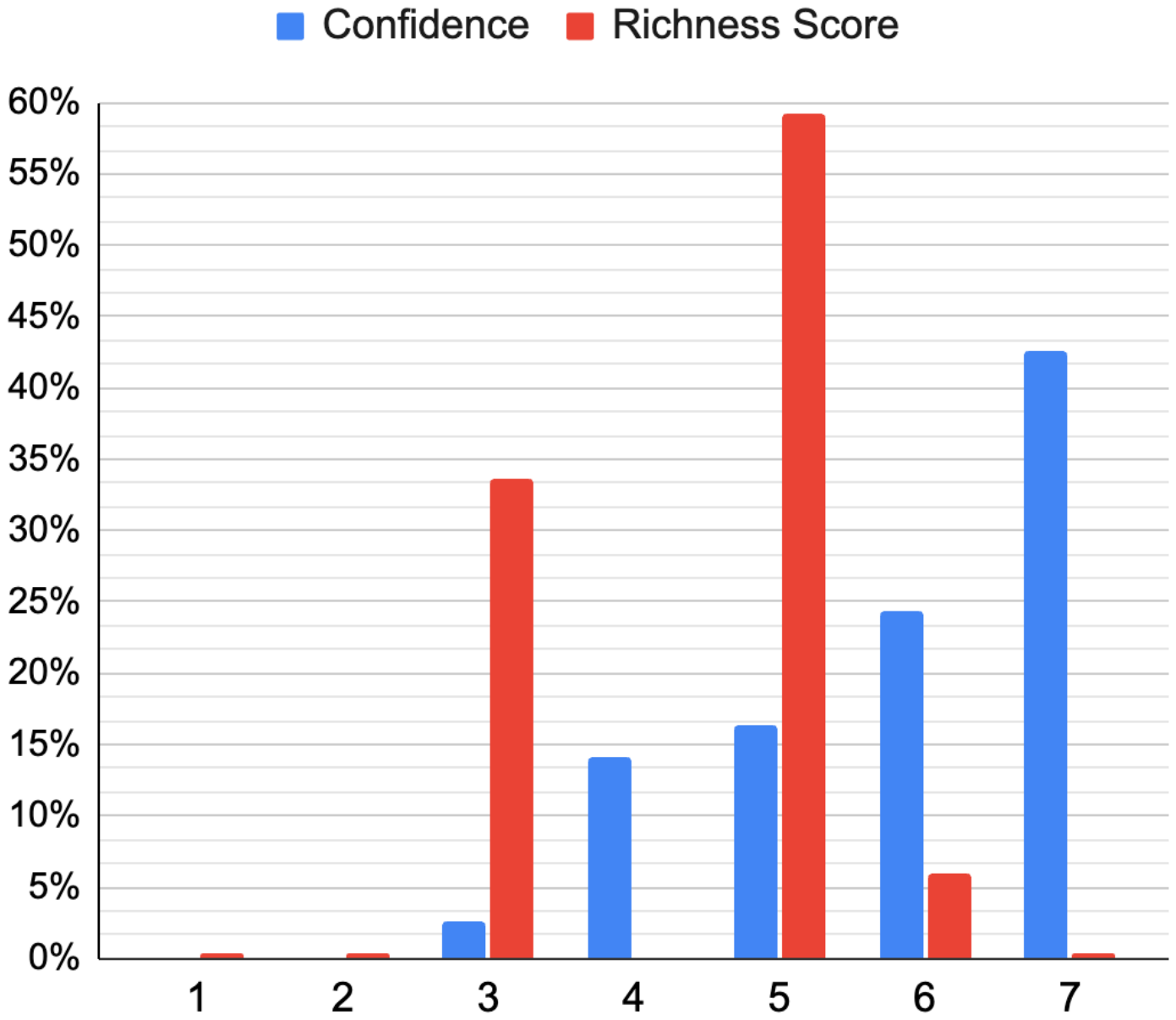}
        \caption{Office Products}
    \end{subfigure}
    \hspace{5mm}
    \begin{subfigure}{0.4\textwidth}
        \includegraphics[width=\linewidth]{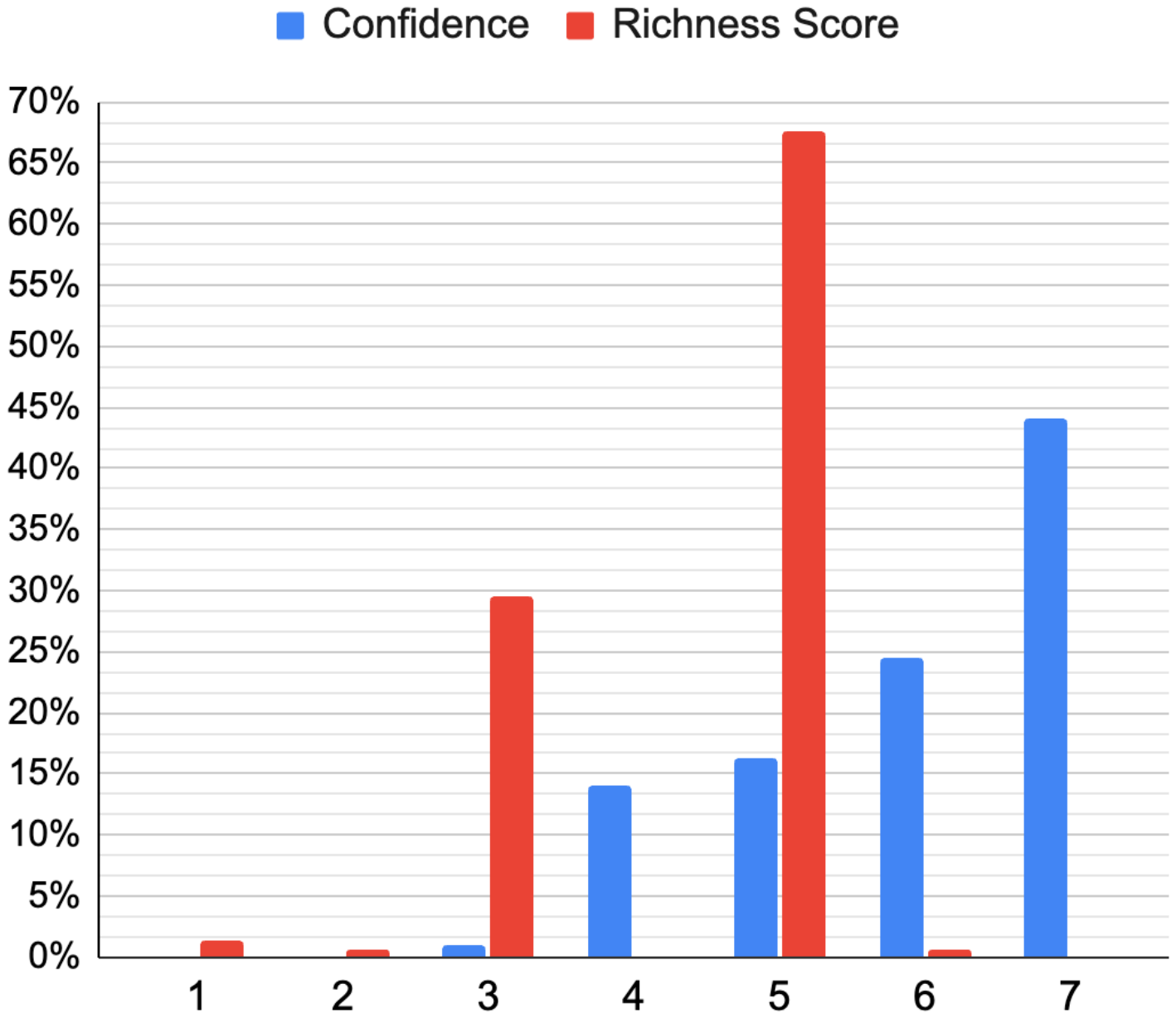}
        \caption{Clothing, Jewelry \& Accessories}
    \end{subfigure}
    \caption{User richness scores and associated confidence.}
    \label{fig:richness_eval}
\end{figure*}
\section{REGEN}

This section details the REGEN dataset, including the data generation process and evaluation methodology with a rater LLM.
\subsection{Amazon Reviews}
We use datasets from Amazon Product Reviews \citep{ni-etal} as a basis for augmentation with rich narratives. The Amazon Reviews dataset (2018) is a massive collection of customer reviews spanning various product categories. This version, released in 2018, contains over 233 million reviews, making it one of the largest publicly available datasets for sentiment analysis and recommendation systems. 
Each review includes item features like title, description, category, price and review features  including text, timestamp, score and summary. User sequences can be created by sorting each user review by timestamp, creating a sequential recommendation task.

For our work, we use ``Office Products'' and ``Clothing, Shoes and Jewelry,'' verticals. These are chosen to be representative samples with different item counts, Office with ~27k items and Clothing with 360k items. We plan to update these to other key verticals, and including the ``All Reviews''.

\subsection{Dataset Generation}

\subsubsection{Data Preprocessing}
User sequences $S_u = \{i_1, i_2, ..., i_n\}$ consisting of interactions $i$ are first created by aggregating reviews per user, and sorting them by timestamps. We then truncate all the sequences to the most recent 50 items and filter items with a missing title.  
This results in ~27K items, 101K users for Office with an average of 7-8 items/user and ~370K items, 1.2m users for Clothing with an average of 9.2 items/user.

\subsubsection{Prompting and Narrative Generation}
Our objective is to create a dataset of natural language narratives that mirrors the diverse interactions within a conversational recommender system. To achieve this, we focus on generating outputs that vary across two key dimensions, 

\begin{itemize}
    \item \textbf{Contextualization:} We generate outputs both with/without explicit contextual information (e.g., user summaries vs explanations of the most recent purchase, endorsement of a recommendation) to understand its influence on the quality and relevance of the generated language.
    \item \textbf{Length:} We further explore generating both short-form and long-form narratives for different conversational scenarios.
\end{itemize}

We use LLMs to generate this data from user reviews, employing carefully designed prompts and iterative evaluation to enhance generation quality and minimize inaccuracies. In this work, we use Gemini 1.5 Flash model, which is a fast and efficient version of the Gemini 1.5 models~\cite{geminiteam2024gemini15unlockingmultimodal}.
We leverage the large token lengths of the LLM to generate natural sounding outputs that process the entire user history $S_u$, for each user $u$. The prompt is fairly straight forward, and uses a task prefix with instructions on the task and the output format, followed by the entire history of the user including the item metadata as well as the reviews, concatenated as text, and finally the instructions on the expected output format (see Appendix~\ref{sec:data_prompt_and_examples} for the prompt). 
We use the entire history of the user along with associated reviews to generate purchase reasons and summaries. However, for product endorsement, we withhold the last user review, so the model learns to endorse a newly recommended item.
Furthermore, we utilize LLMs as automated evaluators to assess the generated outputs across multiple attributes. 
We demonstrate the effectiveness of our approach augmenting the Amazon product review dataset \cite{ni-etal}, and evaluating using the Gemini Pro LLM as the rater. Future work will extend this evaluation to other LLMs and include multiple outputs for each prompt to reduce LM bias and enable more robust benchmarking. Detailed description of individual outputs are included in Appendix~\ref{sec:outputs_attributes}.

\subsection {Evaluation using an Auto Rater}

To evaluate the quality of the generated data, we use Gemini Pro LLM as the rater. We give this rater LLM the task, the history, the generated outputs, and prompt it to score the outputs on different attributes.

\subsubsection{Overall Approach}

We use an ensemble rating process, repeating the rating process several times (10-15) and aggregate these scores to a final score. While this is computationally intensive, we find that it improves the attribute scores with more reliable results. We present distribution of scores across all conversations, and annotate users with corresponding scores in the released dataset. 

\subsubsection{Evaluation Metrics and Confidence Scoring}
We prompt the rater LLM to score the generated data on up to seven attributes (veracity, grounding, clarity \& specificity, foresight, personalization, user richness and confidence) using a seven point likert scale. The definitions are summarized in Appendix~\ref{sec:outputs_attributes}. 
Further, we conduct the evaluation in four stages, prompting the LLM each time:
\begin{itemize}
    \item \textbf{User richness and confidence:} User richness captures the richness of the purchase history for understanding a user.
    \item \textbf{Purchase reason and explanation:} These are scored together based on five key attributes: \textit{veracity, foresight, clarity, personalization, and grounding}. 
    \item  \textbf{User Summaries:} These are assessed on the same attributes as purchase reason, excluding foresight, which is only relevant with contextualized output. 
    \item \textbf{Product endorsements:} These are assessed on the same attributes as the purchase reasons.
\end{itemize}
An associated confidence score is generated for each stage. 
This multi-stage approach is used to avoid any correlation caused by Chain of Thought (CoT) bias across different feature types. 
For ensemble aggregation of individual attribute scores, we use a majority rule, where if over half of the ensemble runs agree on a score, that score becomes the final score. In cases where there's no majority agreement, i.e. with ambiguity, the average score (rounded down) is used.
To determine the confidence scores per feature type, we employ an ensemble method using plurality and averaging. The overall confidence score is calculated by first identifying the most frequent score across all ensemble runs for each attribute. Then, we average these scores across all attributes for that feature type to produce a single, representative confidence score. This approach aims to provide a comprehensive and reliable assessment of the generated data's quality.

\subsubsection{Analysis of Purchase Reasons and Summaries}
Figures~\ref{fig:pre_eval} and \ref{fig:us_eval} show the distribution of rated attributes for purchase reason and user summaries. 
Lower foresight implies less or no post-purchase information is used, while a higher foresight score indicates more or some post-purchase information (i.e, the last review) is incorporated into the reason.  
Analysis of the ``foresight'' scores reveals distinct patterns in purchase behavior between Office Products and Clothing. Office product purchases, often routine and predictable, show a greater reliance on past history (scores 1-4). Conversely, clothing purchases, with their higher item count and potential for evolving trends, emphasize the current context (scores clustered around 4 and above). 
Generated summaries demonstrate slightly higher scores overall, particularly in veracity and personalization, likely due to the model's attention to the complete purchase history.  Associated confidence scores are generally high (4 or above).

Figure \ref{fig:richness_eval} illustrates the user richness and confidence scores. Approximately 34\% of office supply and 28\% of clothing examples fall below average richness, with average or above-average confidence.  The dataset is annotated with attributes rather than filtered to facilitate research on the impact of personalization and data quality on various techniques and methods. We will include the evaluations of product endorsements in a future revision of the paper.

\section{Problem Formulation \& Benchmarking Methodology}
In this section, we describe the experimental setup used to generate narratives. The model architecture employs a multi-stage approach to generate narrative recommendations by integrating collaborative filtering signals and semantic embeddings into an LLM.

\subsection{Task Definitions}
We formalize the conversational recommendation task within the context of user interaction sequences. Consider a dataset comprising sequences $S = \{i_1, i_2, ..., i_n\}$, where each interaction $i \in \mathcal{I}$ is represented as a tuple $(\mathrm{ID}, T, P)$, where $\mathrm{ID}$ is the unique identifier of the item, $T$ is a set of natural language components  $T = \{t^1, t^2, ..., t^m\}$  describing the item. This includes the title, detailed product description, product category, and other relevant textual metadata. $P$ is the personalized text associated with the item in the context of a specific user. This could be a user review, rating, or other user-generated content related to the item.

\begin{figure*}[!htb]
    \centering
    \includegraphics[width=0.75\linewidth]{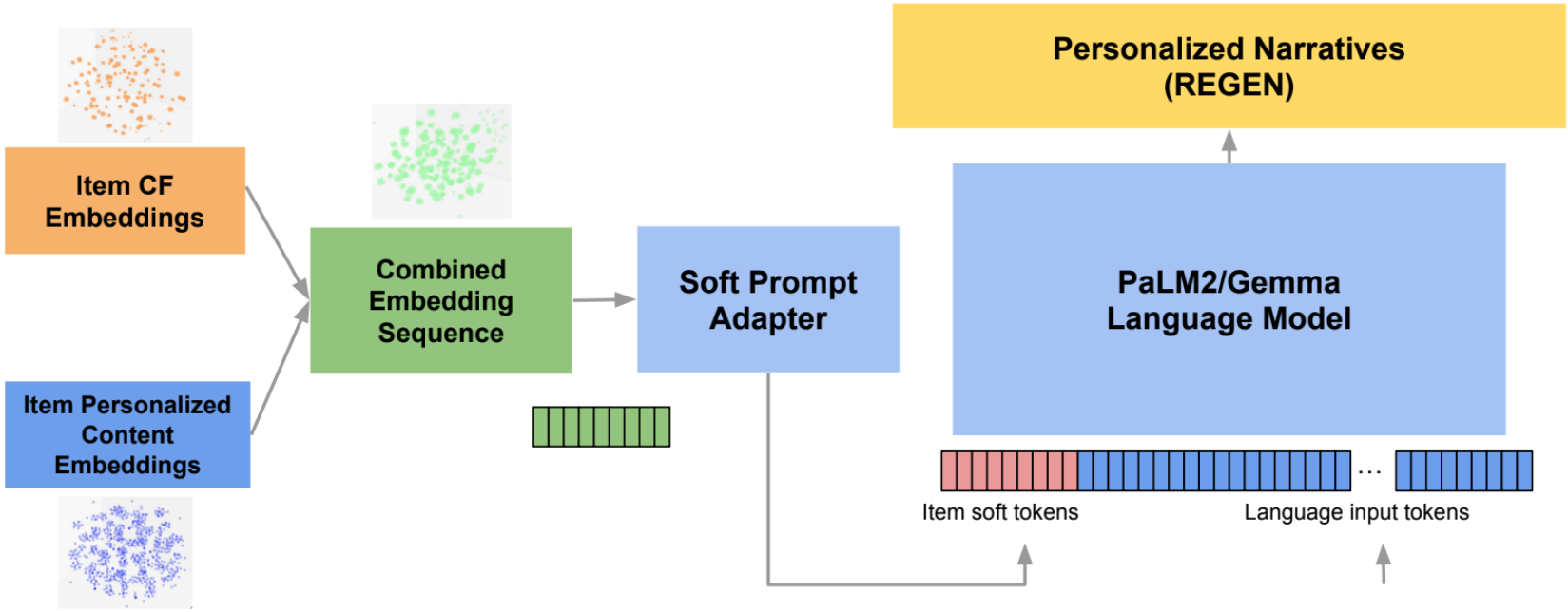}
    \caption{Overview of the Model Architecture.}
    \label{fig:arch}
\end{figure*}

Within this framework, we define three key tasks central to conversational recommendation:

\begin{enumerate}
    \item \textbf{Narrative Generation from Recent Context:} Generate natural language output $\mathcal{T}(i_{n}|S=\{ i_j \mid n - N < j \leq n \})$ corresponding to the recent interaction $i_n$, considering the history of interactions up to and including $n$ (truncated to a certain window of observation $N$). This task focuses on capturing short-term user preferences and generating contextually relevant responses.

    \item \textbf{Narrative Generation from Aggregate Context:} Generate natural language output $\mathcal{T}(S)$ corresponding to the entire interaction history. This task aims to capture long-term user preferences and generate a comprehensive narrative reflecting the user's overall experience.

    \item \textbf{Next Item Prediction:} Predict the next item $i_{n+1}$ in the sequence given the user's past interactions $S$. This task represents the traditional recommendation objective.
\end{enumerate}

In this paper, we focus on the the first two tasks, which are better suited to the proposed dataset, while the third task is evaluated extensively in several recent studies on the reviews data.

\subsection{Model Architecture}

To effectively integrate recommender system knowledge in LLMs for natural language generation, we propose a modular architecture as shown in Figure~\ref{fig:arch} that decouples the encoding of collaborative filtering (CF) signals and semantic representations of user interactions. This approach offers two primary benefits. First, it allows us to systematically investigate the impact of different signal types on the LLM's ability to generate informative and engaging narratives. Second, it serves as a clear and effective baseline for evaluating models trained on the proposed dataset.

Further, by independently selecting best-in-class models for each encoding stage, we can leverage proven methods for generating high-quality item embeddings, such as those from matrix factorization \cite{sysMLmatrix,koren2009matrix}, factorization machines \cite{rendle2010factorization}, or neural collaborative filtering \cite{rendle2020neural}, while simultaneously employing state-of-the-art sentence embedding models like Sentence-T5 \cite{ni2021sentence}, Universal Sentence Encoder \cite{cer2018universal}, GECKO \cite{lee2024geckoversatiletextembeddings}, Sentence-BERT (SBERT) \cite{reimers2019sentence}, SimCSE \cite{gao2021simcse}, and InferSent \cite{conneau2017supervised}. We also note that this modular approach is also highly practical for real-world datasets with item counts orders of magnitude larger (O(1M-1B)) than those typically used in research, enabling efficient encoding of diverse data scales.
Below, we describe the different components used in our multi-stage recommender LM.

\subsubsection{CF Encoder}
This encoder processes the user-item interaction history. For each user $u$, the interaction sequence $S_u = \{i_1, i_2, ..., i_n\}$ consists of interactions $i$, where each interaction is represented as a tuple $(id_i, m_i, s_i, r_i)$ with item ID  $id_i$, item metadata $m_i$, user rating/score for the item $s_i$ and user review for the item $r_i$.

We construct a user-item rating matrix $\mathbf{R}$ from the interaction history and apply Weighted Alternating Least Squares (WALS) \cite{wals4781121} matrix factorization to obtain user embeddings $\mathbf{U} \in \mathbb{R}^{N_u \times d}$ and item embeddings $\mathbf{V} \in \mathbb{R}^{N_i \times d}$, where $N_u$ is the number of users, $N_i$ is the number of items, and $d$ is the embedding dimension of the user and item embeddings. This results in a user embedding $\mathbf{u}_u$ for each user and an item embedding $\mathbf{v}_i$ for each interaction.

\subsubsection{Semantic Encoder}
This captures the semantic information from item metadata and user reviews. For each interaction $i$, we concatenate the item metadata $m_i$ and user review $r_i$ to form a text representation $t_i$. We then apply a pre-trained sentence embedding model (e.g., sT5, Gecko) to $t_i$ to obtain a semantic embedding $\mathbf{s}_i \in \mathbb{R}^{d_s}$, where $d_s$ is embedding size of the sentence embedding model.

\subsubsection{Embedding Fusion}
To combine the collaborative filtering and semantic information, we fuse the embeddings from both encoders.
We note that the semantic embeddings $\mathbf{s}_i$ produced by the Gecko model are each L2 normalized to 1. However the WALS embeddings are not, and the norm of these embeddings is related to the popularity of the item. To keep this intact, we perform a normalization in expected value as $\mathbf{v}_i^{'} = {\mathbf{v}_i}/{\bar{\|\mathbf{v}\|_2}}$, 
where $\bar{\|\mathbf{v}\|_2}$ is the mean L2 norm of all item embeddings.
For each interaction $i$, the item embedding $\mathbf{v}_i^{'}$ and the semantic embedding $\mathbf{s}_i$ are then combined using a simple concatenation, resulting in a combined embedding $\mathbf{c}_i = [\mathbf{v}_i^{'}, \mathbf{s}_i] \in \mathbb{R}^{d_c}$, $d_c = d_i + d_s$. While the model can compensate for the normalization to some extent, we find that this additional preprocessing step improves the metrics.

\subsubsection{Adapter Model}
The sequence of combined embeddings $\mathbf{C}_u = \{\mathbf{c}_1, \mathbf{c}_2, ..., \mathbf{c}_{n}\}$ for user $u$ is fed into into an adapter model (e.g an MLP or a transformer). This adapter model generates a sequence of soft tokens $\mathbf{Z}_u = \{\mathbf{z}_1, \mathbf{z}_2, ..., \mathbf{z}_{n}\}$, where each $\mathbf{z}_i \in \mathbb{R}^{d_l}$ matches the model dimension $d_l$ of the LLM. These soft tokens serve as a condensed representation of the user's interaction history, tailored for integration with the LLM.

\subsubsection{Large Language Model (LLM)}

The soft token sequence $\mathbf{Z}_u$ from the adapter model is prepended to the text prompt token embeddings $\mathbf{P} = \{\mathbf{p}_1, \mathbf{p}_2, ..., \mathbf{p}_{m}\}$ to form the input sequence $\mathbf{I} = \{\mathbf{z}_1, \mathbf{z}_2, ..., \mathbf{z}_{n}, \mathbf{p}_1, \mathbf{p}_2, ..., \mathbf{p}_{m}\}$ for the LLM. This allows the LLM to generate natural language narratives conditioned on both the user's interaction history and the specific prompt, which could indicate additional context. Notably, this just requires $N$ soft tokens to represent $N$ interactions.

\section{Experiments}

For our initial benchmarks, we use i) WALS encoder generating CF embeddings of dimension 128, ii) Gecko 1B encoder generating semantic embeddings of dimension 768 and PaLM 2 XXS~ \cite{anil2023palm2technicalreport} for the LLM, and an MLP layer as the adapter.

\subsection{Datasets and Tasks}

To evaluate the models, we utilize the proposed REGEN dataset, focusing on the task of generating natural language narratives. We demonstrate the model's capabilities as follows:

\paragraph{Narrative Generation} The model takes a user's history of ratings and reviews as input and generates a comprehensive narrative summarizing their preferences and experiences. For the language input, we use a fixed prompt, essentially prompting the model to generate the corresponding output feature based on the observed history.

\paragraph{Evaluation with Distinct Users} The training and test sets contain different users. This forces the model to learn how to interpret user history from the provided embeddings, rather than memorizing specific users. This ensures the model can generalize to new users and accurately capture their preferences through the generated narratives.

We generate benchmarks using REGEN dataset with i) \textbf{Office Products:} with $27K$ items and ii) \textbf{Clothing, Shoes,\& Jewelry:} with $376K$ items, to evaluate the model's performance across both small and large item vocabularies.

\subsection{Evaluation Metrics}

We assess the model using the following complementary methods:

\begin{itemize}
    \item \textbf{Traditional NLP Metrics:} We report BLEU and ROUGE scores to establish baselines. These metrics offer a high-level view of performance trends.
    \item \textbf{Similarity Scores:} We compare the generated narratives to the REGEN refeence narratives using similarity measures based on sentence embedding models (e.g., Gecko). This helps assess the semantic closeness to the ground truth.
    \item \textbf{Side-by-Side Evaluations with raters:} This provides a more nuanced evaluation of the quality and relevance of the generated narratives, either using a human raters or a rater LLM. We do not have comprehensive evaluations using this approach, but show sampled evaluation of some examples SxS with our annotations in Appendix~\ref{sec:sxs_traces}.
\end{itemize}

\subsection{Key Results}
The experiment results are shown in Table~\ref{tab:office}.  

\subsubsection{Natural Language Tasks with Context}
These are tasks where the language outputs are based on immediate context, for example, explaining the reasoning and explanation behind the most recently purchased item. Combining collaborative filtering (CF) and semantic embeddings leads to up to 12\%, +8\%, and +8\% gains in BLEU, ROUGE, and Semantic Similarity metrics, respectively, compared to using the best result using either of these embeddings alone.

\subsubsection{Natural Language Tasks without Context}
For tasks that don't rely on immediate context (e.g., generating a narrative based on an aggregate user profile), combining collaborative filtering and content embeddings offers no significant advantage over using content embeddings alone. This could be because the model does not need to focus on the last item specifically, which is more typical of a recommendation task, and content features are sufficient. 

Overall, we find that the BLEU/ROUGE scores are lower compared to those you'd expect in structured generation tasks. However, our SxS examples in Appendix~\ref{sec:sxs_traces} show that the model outputs compare well to the reference outputs. This is also reflected in the higher similarity scores. 

\begin{table*}
  \centering
  \begin{tabular}{|l|l|c|c|c|c|c|}
    \toprule
    \textbf{Output Feature}  & \textbf{Metrics} & \textbf{CF} & \textbf{Content} & \textbf{CF} + \textbf{Content} & (\%) vs \textbf{CF} & (\%)  vs \textbf{ Content} \\
    \midrule
    Purchase reason   & BLEU &  17.71 & 21.07 & 21.92 & \green{+23.8\%} & \green{+4.0\%} \\
    \rowcolor{lightblue} &   ROUGE-LSum & 38.26 & 42.63 & 43.01 & \green{+12.4\%} & \green{+0.9\%}  \\
    &  Similarity &  0.614 & 0.664 & 0.669 & \green{+9.0\%} & \green{+0.8\%} \\
    Purchase reason explanation   & BLEU & 19.56 & 21.68 & 24.24 & \green{+23.9\%} & \green{11.8\%} \\
    \rowcolor{lightblue} &   ROUGE-LSum &  37.26 & 38.05 & 41.11 & \green{+10.3\%} & \green{+8.0\%} \\
    &  Similarity & 0.783 & 0.787 & 0.851 & \green{+8.7\%} & \green{8.1\%}  \\
    \midrule
    User Summary (Brief)   & BLEU & 17.96 & 19.3 & 19.12 & \green{+6.5\%} & \red{-0.9\%}\\
    \rowcolor{lightblue} &  ROUGE-LSum & 36.5 & 38.06 & 38.16 & \green{+4.5\%} & \green{+0.3\%}  \\
    &  Similarity &  0.861 & 0.866 & 0.864 & \green{+0.3\%} & \red{-0.2\%}\\
    User Summary (Long)   & BLEU & 16.96 & 18.07 & 18.09 & \green{+6.7\%} & \green{+0.1\%} \\
    \rowcolor{lightblue} &  ROUGE-LSum &  31.61 & 33.35 & 33.42 & \green{+5.7\%} & \green{+0.2\%} \\
    &  Similarity & 0.885 & 0.895 & 0.903 & \green{+2.0\%} & \green{+0.9\%} \\
    \bottomrule
  \end{tabular}
  \captionsetup{justification=centering}
  \caption{Performance benchmarks with using different embedding inputs to the PaLM2 XXS LLM. \\ REGEN Office Products ($|I| = 27K$) }
  \label{tab:office}
\end{table*}

\section{Embedding Analysis}
To analyze the learned embeddings, we consider a linear projection layer as the adapter, which yields comparable results (within 5\%) of an MLP adapter.
Let $e_i \in \mathbb{R}^{d_i}$ represent item embedding, $e_s \in \mathbb{R}^{d_s} $ as content embedding, and $W \in \mathbb{R}^{d_l \times (d_i + d_s)} $ as the projection matrix (adapter layer). As we concatenate the item and semantic embeddings, this is equivalent to $W_1 \in \mathbb{R}^{d_l \times d_i}$   submatrix of  $W$ operating on $e_i$ and $W_2 \in \mathbb{R}^{d_l \times d_s}$ submatrix of  $W$ operating on $e_s$. The new embedding can be simply represented as,

\begin{figure}[!htb]
    \centering
    \includegraphics[width=0.75\linewidth]{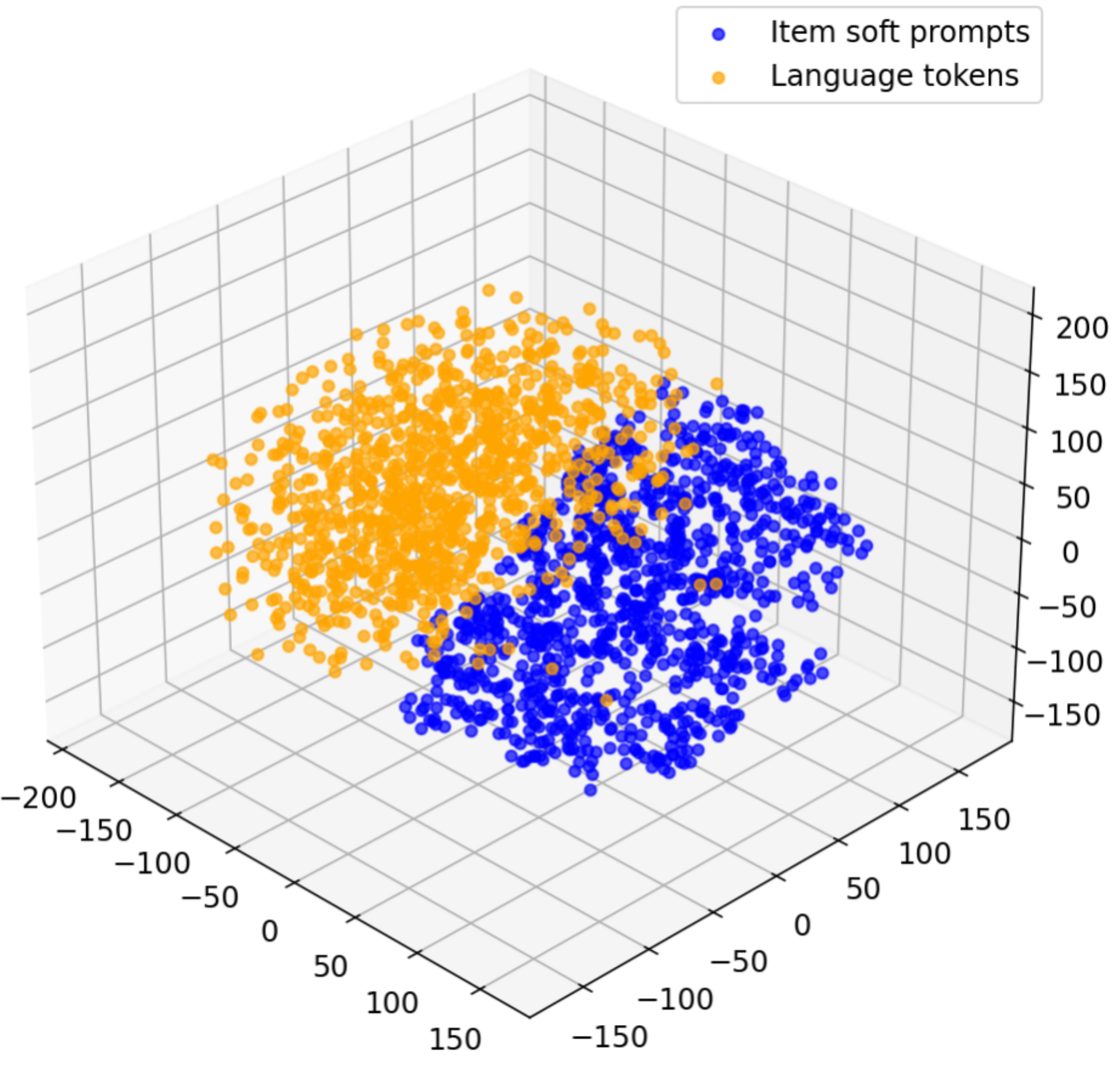}
    \caption{Comparing Soft prompt embeddings vs LLM token embeddings. Soft prompt embeddings generated for all items in Office Products.}
    \label{fig:embeddings}
\end{figure}

\begin{align*}
e_n = W \begin{bmatrix} e_i \\ e_s \end{bmatrix} = \begin{bmatrix} W_1 & W_2 \end{bmatrix} \begin{bmatrix} e_i \\ e_s \end{bmatrix} = W_1 e_i + W_2 e_s
\end{align*}

\begin{figure}[!htb]
    \centering
    \includegraphics[width=0.9\linewidth]{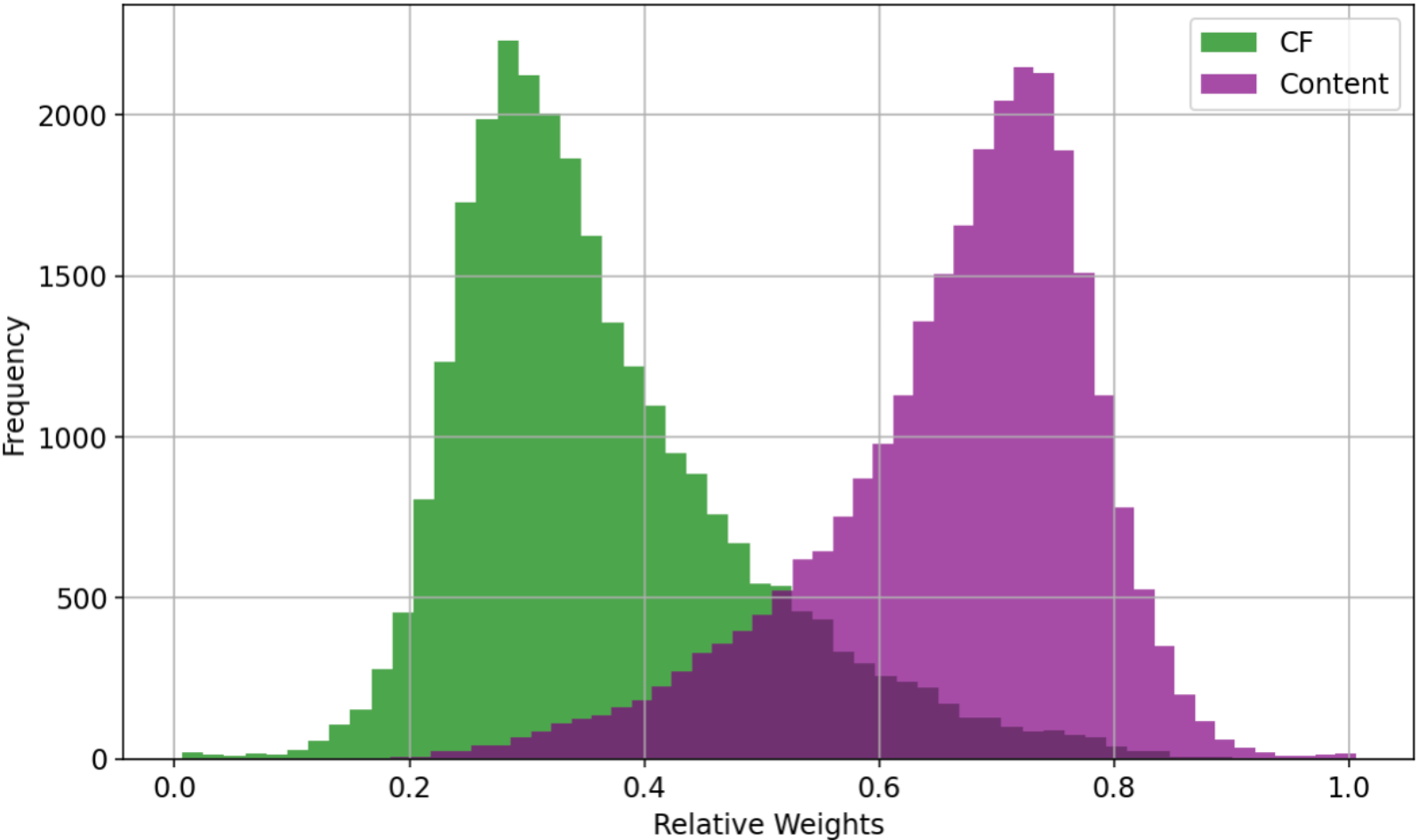}
    \caption{Comparing CF and Content contributions to Soft Prompts (Office Products).}
    \label{fig:relative_weights}
\end{figure}

Essentially, the soft embeddings in the token embedding space are expressed as a projected CF and projected content components, where these projections are learnt by the adapter layer. This framework allows us to get more insight into the contribution of collaborative filtering and content signals.

\subsection{Soft Prompt Embeddings vs LM Token Embeddings:}
We first investigate the relationship between soft prompt embeddings (i.e projected embeddings $e_n$) and the language model's token embeddings. We train a CF + Content model and obtain the projection layer to compute the projected CF and content embeddings for all items. Visualizing these embedding spaces with t-SNE (Figures \ref{fig:embeddings}) reveals that soft prompts, including their constituent CF and content embeddings ($W_1 e_i$, $W_2 e_s$), occupy a distinct space (i.e a different cluster) rather than mapping to the language model's token space. 
While this might seem counter-intuitive at first, it's not a surprising result considering the sparsity of the language embedding space.  The maximum number of embeddings possible in a high-dimensional space with a certain minimum degree of separation is related to the concept of the kissing number in sphere packing \cite{conway1999sphere}.  For instance, in ~25 dimensions, this number is on the order of 100,000 and further increases exponentially with dimensionality. Given that most language models have embedding dimensions in the order of 100s to few 1,000s and vocabulary sizes on the order of 10,000 to 100,000 tokens, the language embeddings utilize a very small fraction of the potential embedding space.  Therefore, it is reasonable that the adapter learns a separate embedding space, especially when trained on smaller datasets and with smaller models. However, it is an open question whether this behavior would persist if recommender language models were pretrained on datasets comparable in scale to those used for general-purpose language models.
Interestingly, this finding also aligns with concepts that treat items as new (discrete) tokens \cite{singh2024bettergeneralizationsemanticids}, effectively expanding the language model's vocabulary, but with the distinction here of utilizing the full unconstrained embedding space representation.
\subsection{CF vs Content Embeddings}
Figure~\ref{fig:relative_weights} shows the distribution of relative contribution (in norm) of projected CF and content embeddings ($W_1 e_i$, $W_2 e_s$) to the total norm of the soft prompt embeddings. This shows that the content based signals have more significant representation in the learnt soft tokens compared to the CF parts, though CF parts are not insignificant. This could explain the observations in our experiments, where we see Content only embedding model outperform the CF based embedding model, while combining the two yields the best results. 
In the Office Products dataset, most items appear in 10-20 users (see Appendix~\ref{sec:item_counts}), which does not give extensive CF signals. 
A similar trend is observed with item statistics in several categories, so it would be informative to repeat these experiments on different datasets with richer interaction data.

\section{Limitations and Future Work}

While we have utilized a state-of-the-art LLM as an automated evaluator for the generated narratives and observed high-quality results across various dimensions, we acknowledge that LLMs cannot fully replace human evaluation, particularly for tasks requiring subjective judgment \cite{chiang2023largelanguagemodelsalternative}. For instance, while LLMs excel at assessing summarization quality, they may struggle with more nuanced aspects like creativity and coherence in narrative generation. Nevertheless, we believe LLMs can effectively complement human evaluation, reducing the need for extensive human involvement and improving efficiency.  Recent work like RoboRater \cite{RoboRater} further supports this notion, demonstrating promising results in automating the evaluation of task-oriented conversations. In future updates of REGEN, we plan to explore augmenting our automated evaluation with human assessments for a more comprehensive evaluation.
For benchmarking, we employed a simple yet effective embedding input-based LLM architecture to incorporate recommender and semantic knowledge. This approach allows for reproducible benchmarks and facilitates the study of how different signal types impact narrative generation. Future work could explore other architectures, such as employing co-training of encoders or techniques like cross-attention and pre-training, to potentially improve performance.

\section{Conclusions}
This paper introduces REGEN, a new dataset designed to advance research in conversational recommender systems. By enhancing the Amazon Product Reviews dataset with richer user narratives, REGEN enables the development and evaluation of models capable of generating personalized explanations and summaries of user preferences.  We highlighted the challenge of bridging the gap between traditional recommender systems and natural language generation, emphasizing the need for innovative methods that effectively combine LLMs with recommender system knowledge.

Through automated evaluation and analysis of generated outputs, we demonstrated REGEN's capacity to capture user preferences and context.  Furthermore, we presented a simple, yet effective fusion architecture that combines collaborative filtering signals and semantic embeddings as input to an LLM, showcasing its potential for generating informative and engaging conversational recommendations. Our experiments demonstrated the effectiveness of this approach in improving natural language generation metrics, particularly when combining both collaborative filtering and semantic information compared to using either of these individually. We also provide an analysis to interpret how CF and content embeddings contribute to this new generative task.

\section{Acknowledgments}
The authors thank James Ren, Ajit Apte, Dima Kuzmin, Walid Krichene who graciously offered their expert advice and feedback on this work. 

\clearpage

\bibliography{main}

\begin{thebibliography}{56}
\providecommand{\natexlab}[1]{#1}

\bibitem[{rec(2022)}]{recsys2022language}
 2022.
\newblock Language Models as Recommender Systems: Evaluations and Limitations.
\newblock In \emph{Proceedings of the 16th ACM Conference on Recommender Systems}. ACM.

\bibitem[{Borgeaud et~al.(2021)Borgeaud, Mensch, Hoffmann, Cai, Rutherford, Millican, Driessche, Lespiau, Bosma, Hall et~al.}]{borgeaud2021improving}
Borgeaud, S.; Mensch, A.; Hoffmann, J.; Cai, T.; Rutherford, E.; Millican, K.; Driessche, G.~v.; Lespiau, J.-B.; Bosma, M.; Hall, J.; et~al. 2021.
\newblock Improving Language Models by Retrieving from Trillions of Tokens.
\newblock \emph{arXiv preprint arXiv:2112.04426}.

\bibitem[{Cer et~al.(2018)Cer, Yang, Kong, Hua, Limtiaco, John, Constant, Guajardo-Cespedes, Yuan, Tar et~al.}]{cer2018universal}
Cer, D.; Yang, Y.; Kong, S.-y.; Hua, N.; Limtiaco, N.; John, R.~S.; Constant, N.; Guajardo-Cespedes, M.; Yuan, S.; Tar, C.; et~al. 2018.
\newblock Universal sentence encoder.
\newblock In \emph{Proceedings of the 2018 Conference on Empirical Methods in Natural Language Processing: System Demonstrations}, 169--174.

\bibitem[{Chen et~al.(2022)Chen, Chen, Ren, Huang, Tang, and Zhao}]{chen2022econvrec}
Chen, L.; Chen, W.; Ren, J.; Huang, F.; Tang, J.; and Zhao, W. 2022.
\newblock E-convrec: A large-scale conversational recommendation dataset for e-commerce customer service.
\newblock In \emph{Proceedings of the ACM Web Conference 2022}, 2164--2175.

\bibitem[{Chen et~al.(2024)Chen, Zuo, Li, Zhang, Mei, and Bendersky}]{chen2024unlockingwhybuyingintroducing}
Chen, T.; Zuo, S.; Li, C.; Zhang, M.; Mei, Q.; and Bendersky, M. 2024.
\newblock Unlocking the `Why' of Buying: Introducing a New Dataset and Benchmark for Purchase Reason and Post-Purchase Experience.
\newblock arXiv:2402.13417.

\bibitem[{Chiang and yi~Lee(2023)}]{chiang2023largelanguagemodelsalternative}
Chiang, C.-H.; and yi~Lee, H. 2023.
\newblock Can Large Language Models Be an Alternative to Human Evaluations?
\newblock arXiv:2305.01937.

\bibitem[{Conneau et~al.(2017)Conneau, Kiela, Schwenk, Barrault, and Bordes}]{conneau2017supervised}
Conneau, A.; Kiela, D.; Schwenk, H.; Barrault, L.; and Bordes, A. 2017.
\newblock Supervised learning of universal sentence representations from natural language inference data.
\newblock In \emph{Proceedings of the 2017 conference on empirical methods in natural language processing}, 670--680.

\bibitem[{Conway and Sloane(1999)}]{conway1999sphere}
Conway, J.; and Sloane, N. 1999.
\newblock \emph{Sphere Packings, Lattices and Groups}.
\newblock Grundlehren der mathematischen Wissenschaften. Springer-Verlag.
\newblock ISBN 9780387985855.

\bibitem[{Doddapaneni et~al.(2024)Doddapaneni, Sayana, Jash, Sodhi, and Kuzmin}]{doddapaneni2024userembeddingmodelpersonalized}
Doddapaneni, S.; Sayana, K.; Jash, A.; Sodhi, S.; and Kuzmin, D. 2024.
\newblock User Embedding Model for Personalized Language Prompting.
\newblock arXiv:2401.04858.

\bibitem[{Gao, Yao, and Chen(2021)}]{gao2021simcse}
Gao, T.; Yao, X.; and Chen, D. 2021.
\newblock Simcse: Simple contrastive learning of sentence embeddings.
\newblock In \emph{Proceedings of the 2021 Conference on Empirical Methods in Natural Language Processing}, 6894--6910.

\bibitem[{Gemini(2024)}]{geminiteam2024gemini15unlockingmultimodal}
Gemini. 2024.
\newblock Gemini 1.5: Unlocking multimodal understanding across millions of tokens of context.
\newblock arXiv:2403.05530.

\bibitem[{Gemma(2024)}]{gemmateam2024gemmaopenmodelsbased}
Gemma. 2024.
\newblock Gemma: Open Models Based on Gemini Research and Technology.
\newblock arXiv:2403.08295.

\bibitem[{Goldman and Vasilevski(2024)}]{RoboRater}
Goldman, S.; and Vasilevski, Y. 2024.
\newblock RoboRater: Automating Ratings of Task-Oriented Conversations using LLMs.
\newblock In \emph{AAAI 2024 Spring Symposium on User-Aligned Assessment of Adaptive AI Systems}.

\bibitem[{Guu et~al.(2020)Guu, Lee, Tung, Chang, and Gao}]{guu2020realm}
Guu, K.; Lee, K.; Tung, Z.; Chang, P.~P.; and Gao, J. 2020.
\newblock REALM: Retrieval-Augmented Language Model Pre-Training.
\newblock In \emph{International Conference on Machine Learning}, 3728--3738. PMLR.

\bibitem[{Hebert et~al.(2024{\natexlab{a}})Hebert, Kyriakidi, Pham, Sayana, Pine, Sodhi, and Jash}]{hebert2024flarefusinglanguagemodels}
Hebert, L.; Kyriakidi, M.; Pham, H.; Sayana, K.; Pine, J.; Sodhi, S.; and Jash, A. 2024{\natexlab{a}}.
\newblock FLARE: Fusing Language Models and Collaborative Architectures for Recommender Enhancement.
\newblock arXiv:2409.11699.

\bibitem[{Hebert et~al.(2024{\natexlab{b}})Hebert, Sayana, Jash, Karatzoglou, Sodhi, Doddapaneni, Cai, and Kuzmin}]{hebert2024persomapersonalizedsoftprompt}
Hebert, L.; Sayana, K.; Jash, A.; Karatzoglou, A.; Sodhi, S.; Doddapaneni, S.; Cai, Y.; and Kuzmin, D. 2024{\natexlab{b}}.
\newblock PERSOMA: PERsonalized SOft ProMpt Adapter Architecture for Personalized Language Prompting.
\newblock arXiv:2408.00960.

\bibitem[{Hu, Koren, and Volinsky(2008)}]{wals4781121}
Hu, Y.; Koren, Y.; and Volinsky, C. 2008.
\newblock Collaborative Filtering for Implicit Feedback Datasets.
\newblock In \emph{2008 Eighth IEEE International Conference on Data Mining}, 263--272.

\bibitem[{Izacard et~al.(2022)Izacard, Caron, Hosseini, Riedel, Boissonnade, Joulin, and Grave}]{izacard2022atlas}
Izacard, G.; Caron, M.; Hosseini, A.; Riedel, S.; Boissonnade, G.; Joulin, A.; and Grave, E. 2022.
\newblock Atlas: Few-shot Learning with Retrieval Augmented Language Models.
\newblock In \emph{International Conference on Learning Representations}.

\bibitem[{Izacard and Grave(2021)}]{izacard2021leveraging}
Izacard, G.; and Grave, E. 2021.
\newblock Leveraging Passage Retrieval with Generative Models for Open Domain Question Answering.
\newblock In \emph{International Conference on Learning Representations}.

\bibitem[{Jannach et~al.(2020)Jannach, Kumar, Lerche, and Zanker}]{jannach2020survey}
Jannach, D.; Kumar, A.; Lerche, L.; and Zanker, M. 2020.
\newblock A survey on conversational recommender systems.
\newblock In \emph{Proceedings of the 25th ACM SIGKDD International Conference on Knowledge Discovery \& Data Mining}, 3637--3638.

\bibitem[{Kim et~al.(2024)Kim, Kang, Choi, Kim, Yang, and Park}]{kddLLMCF}
Kim, S.; Kang, H.; Choi, S.; Kim, D.; Yang, M.; and Park, C. 2024.
\newblock Large Language Models meet Collaborative Filtering: An Efficient All-round LLM-based Recommender System.
\newblock In \emph{Proceedings of the 30th ACM SIGKDD Conference on Knowledge Discovery and Data Mining}, KDD '24, 1395–1406. New York, NY, USA: Association for Computing Machinery.
\newblock ISBN 9798400704901.

\bibitem[{Koren, Bell, and Volinsky(2009)}]{koren2009matrix}
Koren, Y.; Bell, R.; and Volinsky, C. 2009.
\newblock Matrix factorization techniques for recommender systems.
\newblock \emph{Computer}, (8): 30--37.

\bibitem[{Lee et~al.(2024)Lee, Dai, Ren, Chen, Cer, Cole, Hui, Boratko, Kapadia, Ding, Luan, Duddu, Abrego, Shi, Gupta, Kusupati, Jain, Jonnalagadda, Chang, and Naim}]{lee2024geckoversatiletextembeddings}
Lee, J.; Dai, Z.; Ren, X.; Chen, B.; Cer, D.; Cole, J.~R.; Hui, K.; Boratko, M.; Kapadia, R.; Ding, W.; Luan, Y.; Duddu, S. M.~K.; Abrego, G.~H.; Shi, W.; Gupta, N.; Kusupati, A.; Jain, P.; Jonnalagadda, S.~R.; Chang, M.-W.; and Naim, I. 2024.
\newblock Gecko: Versatile Text Embeddings Distilled from Large Language Models.
\newblock arXiv:2403.20327.

\bibitem[{Lester, Al-Rfou, and Constant(2021)}]{lester2021powerscaleparameterefficientprompt}
Lester, B.; Al-Rfou, R.; and Constant, N. 2021.
\newblock The Power of Scale for Parameter-Efficient Prompt Tuning.
\newblock arXiv:2104.08691.

\bibitem[{Lewis et~al.(2020)Lewis, Perez, Piktus, Petroni, Karpukhin, Goyal, Küttler, Zettlemoyer, Devlin, Chen et~al.}]{lewis2020retrieval}
Lewis, P.; Perez, E.; Piktus, A.; Petroni, F.; Karpukhin, V.; Goyal, N.; Küttler, H.; Zettlemoyer, L.; Devlin, J.; Chen, W.-t.~Y.; et~al. 2020.
\newblock Retrieval-Augmented Generation for Knowledge-Intensive NLP Tasks.
\newblock \emph{Advances in Neural Information Processing Systems}, 33: 19690--19701.

\bibitem[{Li et~al.(2017)Li, Zhao, He, Zhang, Wu, and Chang}]{li2017redial}
Li, X.; Zhao, L.; He, X.; Zhang, M.; Wu, Q.; and Chang, P.~P. 2017.
\newblock Redial: A large-scale dataset for group recommendation.
\newblock In \emph{Proceedings of the 40th international ACM SIGIR conference on research and development in information retrieval}, 1157--1160.

\bibitem[{Li and Liang(2021)}]{li2021prefixtuningoptimizingcontinuousprompts}
Li, X.~L.; and Liang, P. 2021.
\newblock Prefix-Tuning: Optimizing Continuous Prompts for Generation.
\newblock arXiv:2101.00190.

\bibitem[{Li et~al.(2024)Li, Zhai, Alzantot, Yu, Vulić, Korhonen, and Hammad}]{calrec24}
Li, Y.; Zhai, X.; Alzantot, M.; Yu, K.; Vulić, I.; Korhonen, A.; and Hammad, M. 2024.
\newblock CALRec: Contrastive Alignment of Generative LLMs for Sequential Recommendation.

\bibitem[{Lin et~al.(2021)Lin, Lee, Sun, Hwang et~al.}]{lin2021m6}
Lin, C.-T.; Lee, H.-Y.; Sun, M.-F.; Hwang, W.-C.; et~al. 2021.
\newblock M6-Rec: Generative Pretrained Language Models for Conversational Recommendation.
\newblock In \emph{Proceedings of the 15th ACM Conference on Recommender Systems}, 629--637.

\bibitem[{Lin et~al.(2024)Lin, Wang, Li, Feng, Ng, and Chua}]{kddBridgingItemsLanguage}
Lin, X.; Wang, W.; Li, Y.; Feng, F.; Ng, S.-K.; and Chua, T.-S. 2024.
\newblock Bridging Items and Language: A Transition Paradigm for Large Language Model-Based Recommendation.
\newblock In \emph{Proceedings of the 30th ACM SIGKDD Conference on Knowledge Discovery and Data Mining}, KDD '24, 1816–1826. New York, NY, USA: Association for Computing Machinery.
\newblock ISBN 9798400704901.

\bibitem[{Liu et~al.(2022{\natexlab{a}})Liu, Lian, Fu, Wang, and Xie}]{liu2022carca}
Liu, D.; Lian, J.; Fu, Z.; Wang, Z.; and Xie, X. 2022{\natexlab{a}}.
\newblock CARCA: Content-aware representation for cross-attention based sequential recommendation.
\newblock \emph{Proceedings of the ACM Web Conference 2022}, 1195--1203.

\bibitem[{Liu et~al.(2022{\natexlab{b}})Liu, Huang, Tang, Chen, Zhu, Zhu, and Xie}]{liu2022durecdial}
Liu, Z.; Huang, J.; Tang, J.; Chen, Y.; Zhu, J.; Zhu, X.; and Xie, X. 2022{\natexlab{b}}.
\newblock DuRecDial: A DuReader-based dataset for conversational recommendation.
\newblock In \emph{Findings of the Association for Computational Linguistics: NAACL 2022}, 2832--2842.

\bibitem[{Longpre et~al.(2024)Longpre, Suri, Nguyen, Li, Dai, and Zhao}]{longpre2024foundational}
Longpre, S.; Suri, A.; Nguyen, T.~M.; Li, S.~H.; Dai, A.~M.; and Zhao, Z. 2024.
\newblock Foundational Autoraters: Taming Large Language Models for Better Automatic Evaluation.
\newblock \emph{arXiv preprint arXiv:2407.10817}.

\bibitem[{Nakano et~al.(2021)Nakano, Hilton, Balaji, Jain, and Schulman}]{nakano2021webgpt}
Nakano, R.; Hilton, J.; Balaji, S.; Jain, J.; and Schulman, J. 2021.
\newblock WebGPT: Browser-assisted question-answering with human feedback.
\newblock \emph{arXiv preprint arXiv:2112.09332}.

\bibitem[{Ni et~al.(2021)Ni, Constant, Guajardo-Cespedes, and Gao}]{ni2021sentence}
Ni, J.; Constant, N.; Guajardo-Cespedes, M.; and Gao, J. 2021.
\newblock Sentence-t5: Scalable sentence encoders from pre-trained text-to-text models.
\newblock In \emph{Proceedings of the 2021 Conference on Empirical Methods in Natural Language Processing}, 8133--8142.

\bibitem[{Ni, Li, and McAuley(2019)}]{ni-etal}
Ni, J.; Li, J.; and McAuley, J. 2019.
\newblock Justifying Recommendations using Distantly-Labeled Reviews and Fine-Grained Aspects.
\newblock In \emph{Proceedings of the 2019 Conference on Empirical Methods in Natural Language Processing and the 9th International Joint Conference on Natural Language Processing (EMNLP-IJCNLP)}, 188--197. Hong Kong, China: Association for Computational Linguistics.

\bibitem[{Ning et~al.(2024)Ning, Liu, Wu, Wu, Berlowitz, Prakash, Green, O'Banion, and Xie}]{ning2024userllmefficientllmcontextualization}
Ning, L.; Liu, L.; Wu, J.; Wu, N.; Berlowitz, D.; Prakash, S.; Green, B.; O'Banion, S.; and Xie, J. 2024.
\newblock User-LLM: Efficient LLM Contextualization with User Embeddings.
\newblock arXiv:2402.13598.

\bibitem[{PaLM2(2023)}]{anil2023palm2technicalreport}
PaLM2. 2023.
\newblock PaLM 2 Technical Report.
\newblock arXiv:2305.10403.

\bibitem[{Qi et~al.(2020)Qi, Zhang, Huang, Wu, Zhang, and Zhao}]{qi2020user}
Qi, W.; Zhang, Y.; Huang, Y.; Wu, C.; Zhang, J.; and Zhao, W. 2020.
\newblock U-BERT: Pre-training user representations for improved recommendation.
\newblock In \emph{Proceedings of the 29th ACM International Conference on Information \& Knowledge Management}, 2069--2078.

\bibitem[{Qian et~al.(2021)Qian, Chen, Li, Lin, Huang, and Yang}]{qian2021cafe}
Qian, Z.; Chen, W.-L.; Li, X.; Lin, B.; Huang, X.; and Yang, H. 2021.
\newblock CaFe: Coarse-to-Fine Embedding for Sequential Recommendation.
\newblock In \emph{Proceedings of the 27th ACM SIGKDD Conference on Knowledge Discovery \& Data Mining}, 3032--3040.

\bibitem[{Raffel et~al.(2020)Raffel, Shazeer, Roberts, Lee, Narang, Matena, Zhou, Li, and Liu}]{raffel2020exploring}
Raffel, C.; Shazeer, N.; Roberts, A.; Lee, K.; Narang, S.; Matena, M.; Zhou, Y.; Li, W.; and Liu, P.~J. 2020.
\newblock Exploring the limits of transfer learning with a unified text-to-text transformer.
\newblock \emph{The Journal of Machine Learning Research}, 21(1): 5485--5551.

\bibitem[{Reimers and Gurevych(2019)}]{reimers2019sentence}
Reimers, N.; and Gurevych, I. 2019.
\newblock Sentence-bert: Sentence embeddings using siamese bert-networks.
\newblock In \emph{Proceedings of the 2019 Conference on Empirical Methods in Natural Language Processing and the 9th International Joint Conference on Natural Language Processing (EMNLP-IJCNLP)}, 3982--3992.

\bibitem[{Rendle(2010)}]{rendle2010factorization}
Rendle, S. 2010.
\newblock Factorization machines.
\newblock In \emph{2010 IEEE International Conference on Data Mining}, 995--1000. IEEE.

\bibitem[{Rendle et~al.(2020)Rendle, Krichene, Zhang, and Anderson}]{rendle2020neural}
Rendle, S.; Krichene, W.; Zhang, L.; and Anderson, J. 2020.
\newblock Neural collaborative filtering vs. matrix factorization revisited.
\newblock In \emph{Fourteenth ACM Conference on Recommender Systems}, 240--248.

\bibitem[{Sayana et~al.(2024)Sayana, Vasudeva, Vasilevski, Pine, Hebert, and Pham}]{regendata}
Sayana, K.; Vasudeva, R.; Vasilevski, Y.; Pine, J.; Hebert, L.; and Pham, H. 2024.
\newblock REGEN: Reviews Enhanced with Generative Narratives.

\bibitem[{Singh et~al.(2024)Singh, Vu, Mehta, Keshavan, Sathiamoorthy, Zheng, Hong, Heldt, Wei, Tandon, Chi, and Yi}]{singh2024bettergeneralizationsemanticids}
Singh, A.; Vu, T.; Mehta, N.; Keshavan, R.; Sathiamoorthy, M.; Zheng, Y.; Hong, L.; Heldt, L.; Wei, L.; Tandon, D.; Chi, E.~H.; and Yi, X. 2024.
\newblock Better Generalization with Semantic IDs: A Case Study in Ranking for Recommendations.
\newblock arXiv:2306.08121.

\bibitem[{Sun et~al.(2020)Sun, Liu, Wu, Pei, Lin, Ou, and Jiang}]{sun2020bert4rec}
Sun, F.; Liu, J.; Wu, J.; Pei, C.; Lin, X.; Ou, W.; and Jiang, P. 2020.
\newblock BERT4Rec: Sequential recommendation with bidirectional encoder representations from transformer.
\newblock \emph{Proceedings of the ACM SIGKDD international conference on knowledge discovery and data mining}, 1441--1450.

\bibitem[{Tennenholtz et~al.(2024)Tennenholtz, Chow, Hsu, Shani, Liang, and Boutilier}]{tennenholtz2024embeddingalignedlanguagemodels}
Tennenholtz, G.; Chow, Y.; Hsu, C.-W.; Shani, L.; Liang, E.; and Boutilier, C. 2024.
\newblock Embedding-Aligned Language Models.
\newblock arXiv:2406.00024.

\bibitem[{Thoppilan et~al.(2022)Thoppilan, De~Freitas, Bradley, Chintagunta, Collins, Fedus, Huang, Joulin, Kappler, Keene et~al.}]{thoppilan2022lamda}
Thoppilan, R.; De~Freitas, D.; Bradley, J.; Chintagunta, S.; Collins, M.; Fedus, W.; Huang, M.; Joulin, A.; Kappler, M.; Keene, S.; et~al. 2022.
\newblock LaMDA: Language Models for Dialog Applications.
\newblock \emph{arXiv preprint arXiv:2201.08239}.

\bibitem[{Wu et~al.(2020)Wu, Cao, Wu, Hu, Lian, Xie, Chen, and Liu}]{wu2020mind}
Wu, F.; Cao, Y.~Q.; Wu, L.; Hu, C.; Lian, J.; Xie, Y.; Chen, Y.; and Liu, H. 2020.
\newblock MIND: A large-scale dataset for news recommendation.
\newblock In \emph{Proceedings of the 43rd International ACM SIGIR conference on research and development in information retrieval}, 2431--2434.

\bibitem[{Wu et~al.(2021)Wu, Liang, Chen, Yuan, and Zhang}]{wu2021s3rec}
Wu, K.; Liang, W.~X.; Chen, L.; Yuan, B.; and Zhang, S.~Y. 2021.
\newblock S3-rec: Self-supervised learning for sequential recommendation with mutual information maximization.
\newblock \emph{Proceedings of the 44th International ACM SIGIR Conference on Research and Development in Information Retrieval}, 1262--1271.

\bibitem[{Yang et~al.(2024)Yang, Subbiah, Patel, Li, Song, Mirghaderi, and Aggarwal}]{ilm}
Yang, L.; Subbiah, A.; Patel, H.; Li, J.; Song, Y.; Mirghaderi, R.; and Aggarwal, V. 2024.
\newblock Item-Language Model for Conversational Recommendation.

\bibitem[{Yi et~al.(2018)Yi, Chen, Ramesh, Rajashekhar, Hong, Fiedel, Seshadri, Heldt, Wu, and Chi}]{sysMLmatrix}
Yi, X.; Chen, Y.-F.; Ramesh, S.; Rajashekhar, V.; Hong, L.; Fiedel, N.; Seshadri, N.; Heldt, L.; Wu, X.; and Chi, E.~H. 2018.
\newblock Factorized Deep Retrieval and Distributed TensorFlow Serving.
\newblock \emph{SysML}.

\bibitem[{Yuan et~al.(2022)Yuan, Yang, He, Karatzoglou, and Zhang}]{yuan2022tiger}
Yuan, F.; Yang, X.; He, X.; Karatzoglou, A.; and Zhang, L. 2022.
\newblock TIGER: Transferable interactive graph-enhanced recommender.
\newblock In \emph{Proceedings of the ACM Web Conference 2022}, 398--407.

\bibitem[{Zhang et~al.(2021)Zhang, Sun, Zhang, Zhang, Chen, and He}]{zhang2021topic}
Zhang, Y.; Sun, Y.; Zhang, W.; Zhang, Y.; Chen, X.; and He, X. 2021.
\newblock Topic-guided conversational recommendation.
\newblock In \emph{Proceedings of the 44th International ACM SIGIR Conference on Research and Development in Information Retrieval}, 450--459.

\bibitem[{Zhang et~al.(2022)Zhang, Wu, Zhang, Sun, and Zhang}]{zhang2022generating}
Zhang, Y.; Wu, C.; Zhang, L.; Sun, Z.; and Zhang, W. 2022.
\newblock Generating Personalized Recommendations with Large Language Models.
\newblock \emph{arXiv preprint arXiv:2206.01407}.

\end{thebibliography}

\appendix

\section{Reproducibility}
The REGEN dataset presented in this paper is available for public use~\cite{regendata}. We have used WALS algorithm, which is well known for generating collaborative filtering embeddings. For content embedding, we use Gecko 1B embeddings, which are again publicly available. For benchmarks, we have used a PaLM2 XXS model that can be run with reasonable compute to generate the benchmarks. We also plan to update this with results from a Gemma model~\cite{gemmateam2024gemmaopenmodelsbased}, which is also available for public use.

\section{Details of narrative outputs and eval attributes} \label{sec:outputs_attributes}
The detailed descriptions of narrative outputs and eval attributes are provided in Tables~\ref{tab:nloutputs} and \ref{tab:rating_dimensions}.
\begin{table*}[!htb]
\centering
\begin{tabular}{ |m{4cm}|>{\em}m{6cm}<{\normalfont}| } 
 \hline
 \textbf{Generated output} & \textbf{Description} \\
 \hline
 Product Endorsement & A tailored product endorsement/sales pitch crafted based on understanding user's unique purchase history and reviews  \\
 \hline
 Purchase Reasons & Concise explanations of the reasons behind a product recommendation.  \\ 
 \hline
 Purchase Reason Explanations & Detailed and elaborate explanations of a purchase decision of the most recent item in context of the entire history.  \\ 
 \hline
 Brief User Summaries & Concise summaries of a user's preferences and purchase history.  \\ 
 \hline
 Detailed User Summaries & Comprehensive summaries of a user's preferences and history, potentially including specific examples and justifications.  \\ 
 \hline
 User Profiles & A short phrase describing the type of user in product/shopping context.  \\
 \hline
\end{tabular}
\caption{Generated outputs and their descriptions. Average length is mean number of words for each output.}
\label{tab:nloutputs}
\end{table*}

\begin{table*}[!htb]
\centering
\begin{tabular}{ |m{3cm}|>{\em}m{8cm}<{\normalfont}| } 
\hline
\textbf{\begin{tabular}{l} Rating \\ Attribute \end{tabular}} & \textbf{Description} \\ 
\hline
Veracity &  Accuracy and truthfulness of the purchase reason, analyzing specific evidence and inconsistencies within the user's purchase history. \\
\hline
Grounding & Strength of the supporting evidence for all individual claims made in the purchase reason and explanation. \\
\hline
\begin{tabular}{l} Clarity \& \\ Specificity \end{tabular} &  Clarity and specificity of the purchase reason and its explanation, and specificity of evidence. \\
\hline
Foresight &  How much of the information used in the purchase reason came from the post-purchase review of the last purchased item. \\
\hline
Personalization \ & (User needs vs Product descriptions) Extent to which the narrative prioritizes information directly from the user's reviews and statements, as opposed to relying on product descriptions or general assumptions. \\
\hline
User richness & How information-rich the user's purchase and review history is for understanding their motivations and preferences. \\
\hline
Confidence & Confidence level in the evaluation, ranging from unreliable to high agreement between different evaluation methods. \\
\hline
\end{tabular}
\caption{Auto Rater: Rating Attributes.}
\label{tab:rating_dimensions}
\end{table*}

\section{Office vs Clothing}\label{sec:office_vs_clothing}
In the Clothing category (Table~\ref{tab:clothing}), we observe similar trends with gains when using the combined approach. Interestingly, the metrics for CF embeddings alone are much closer to those of the content-based approach. This suggests that the larger dataset, with increased diversity of items, may be contributing to the stronger performance of collaborative filtering in this category. 

\begin{table*}
  \centering
  \begin{tabular}{|l|l|c|c|c|c|c|}
    \toprule
    \textbf{Output Feature}  & \textbf{Metrics} & \textbf{CF} & \textbf{Content} & \textbf{CF} + \textbf{Content} & (\%) vs \textbf{  CF} & (\%)  vs \textbf{ Content} \\
    \midrule
    Purchase reason explanation   & BLEU & 17.81 & 18.01 & 18.96 & \green{+6.5\%} & \green{+5.3\%}  \\
    \rowcolor{lightblue} &   ROUGE & 34.05 &  34.73 & 35.67 & \green{+4.8\%} & \green{+2.7\%}   \\
    &  Similarity & 0.795 & 0.82 & 0.838 & \green{+5.4\%} & \green{+2.2\%}  \\ 
    \midrule
    User Summary (Long)   & BLEU &  19.18 & 18.92 & 19.49 & \green{+1.6\%} & \green{+3.0\%} \\
    \rowcolor{lightblue} &  ROUGE &  33.01 & 33.23 & 33.88 & \green{+2.6\%} & \green{+2.0\%} \\
    &  Similarity &  0.908 & 0.91 & 0.913 & \green{+0.6\%} & \green{+0.3\%}\\
    \bottomrule
  \end{tabular}
  \captionsetup{justification=centering}
  \caption{Performance benchmarks with using different embedding inputs to the PaLM2 XXS LLM. \\REGEN Clothing, Shoes \& Jewelry ($|I| = 376K$) }
  \label{tab:clothing}
\end{table*}

\section{Item Count distribution}\label{sec:item_counts}
\begin{figure}[!htb]
    \centering
    \includegraphics[width=0.9\linewidth]{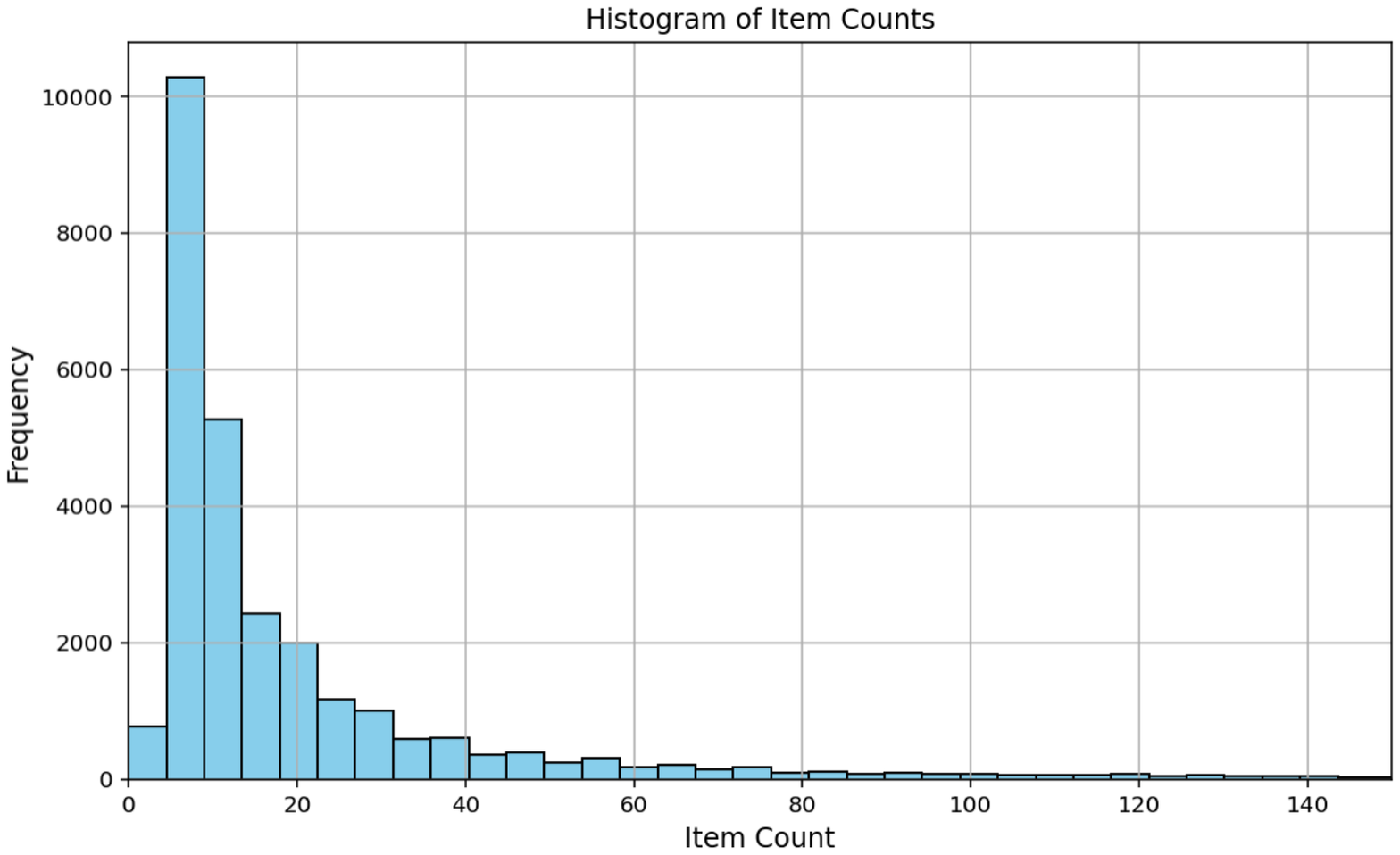}
    \caption{Item count distribution for Office Products.}
    \label{fig:office_items}
\end{figure}

Item count distribution for Office Products is shown in Figure~\ref{fig:office_items} with significant weight at 10-20 item counts.

\section{Comparing CF, Content and Language Embeddings}\label{sec:cf_content_lm_embeddings}
\begin{figure}[!htb]
    \centering
    \includegraphics[width=0.9\linewidth]{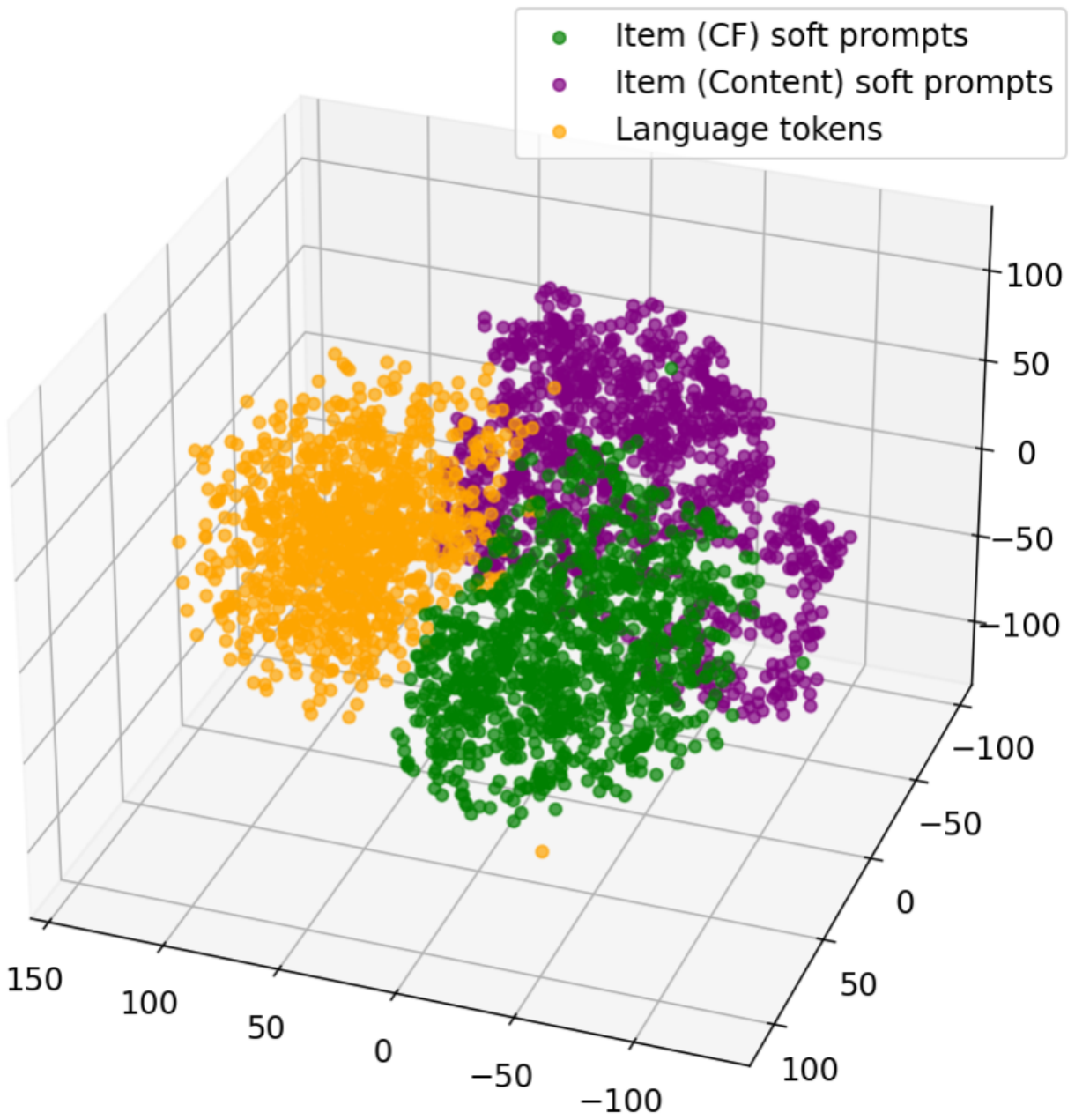}
    \caption{Comparing Soft prompt embeddings vs LLM token embeddings. Soft prompt embeddings generated for all items in Office Products.}
    \label{fig:embeddings_v2}
\end{figure}

Figure~\ref{fig:embeddings_v2} illustrates CF, content embedding components of the soft prompts individually, along with the LM token embeddings plotted with tSNE.

\section{Data generation prompt and Examples} \label{sec:data_prompt_and_examples}
The data generation prompts are shown in Tables~\ref{tab:prompt} and \ref{tab:data_example}.

\section{Auto Rater Prompts} \label{sec:rater_prompts}
An example of auto rater prompt is show in Table~\ref{tab:rater_prompt}.

\section{Example traces} \label{sec:sxs_traces}
These traces are generated from the trained CF+Content embedding model with PaLM2 XXS as the LLM. Gecko embeddings are used to represent content. The distilled embedding LM demonstrates strong performance in generating concise narratives that capture the core user preferences. However, compared to the reference model, it may sometimes oversimplify complex scenarios or miss subtle details.  Furthermore, a tendency to hallucinate information is sometimes observed, especially with longer outputs. Full user histories were used for SxS comparison, but are omitted here for brevity. 
Traces generated by the models are shown in Tables \ref{tab:pre_traces} and \ref{tab:us_traces}.


\onecolumn
\begin{longtable}{|p{0.9\textwidth}|} 
\caption{Prompts used for Data Generation}
\label{tab:prompt}
\\
\hline 
\\
\textit{\textbf{NOTE}: Few shot prompting is used for purchase reasons and user summaries, and a zero shot prompt for product endorsements.} \\
\\
\textbf{Prompt for purchase reasons and user summaries:} \\
You are an exceptional customer engagement specialist at Amazon. Your role is to understand customer purchase motivations and experiences. \\
You will be provided with a sequence of product purchases made by a user and their corresponding reviews. \\
Each sequence is ordered chronologically, from earliest to most recent purchase, marked by 'Sequence Item Position'. \\
\\
Your task is two fold. \\
- One, focus on the last product in their purchase sequence (marked as 'Final Item: True'). Analyze the final purchase based on their prior purchases and reviews. \\
- Two, identify purchase attributes of this customer's purchases using all items in purchase and review history. \\
\\
Provide the following information in JSON format: \\

1. **purchase reason**: The reason for purchase of the last product that can be inferred from the review, product description, and the user's purchase history. If not mentioned, leave null. \\
2. **purchase reason explanation**: Briefly justify your reasoning for the identified purchase reasons, considering the user's past purchases and reviews. provide an in depth user summary based on specific shopping interests \\
3. **brief user summary**: Based on this user's purchase history and reviews, provide  a brief two sentence user summary based on specific shopping interests, including any specific items, brands, colors and attributes. \\
4. **long user summary**: Based on this user's purchase history and reviews, provide an in depth user summary based on specific shopping interests, including any specific items, brands, colors and attributes. \\
5. **User profile**: Provide a user profile in 4 to 6 words based on the user's past purchases and reviews. \\
\\
\textbf{Example JSON Output:}\\
\\
json \\
\{ \\
  "purchase reason": "........", \\
  "purchase reason explanation": "........", \\
  "brief user summary": "........", \\
  "long user summary": "........", \\
  "user profile": "........" \\
\} \\
\\

*Important:** \\
**DO NOT** quote  any part of the user's reviews verbatim in your answers.\\
Focus on insights, not direct repetition.\\
Be specific and relate to the context of each customer's purchase and review history. \\

\\
Here are examples of 7 reviewers and the expected answer for the last item in the sequence for each reviewer id\:
\\
\\
{\color{darkgreen}reviewer id\: A01003458IEUPS8LQ1QU }\\
{\color{darkgreen}Sequence Item Position\: 1 }\\
Final Item\: False \\
        item\_title\: Rolodex Open Rotary Business Card File \\
        item\_desc\: Unique Stack-Lock feature keeps stacked trays together with locked-in stability Unique Stack-Lock feature keeps stacked trays together with locked-in stability \\
        item\_price\: 9.989999771118164 \\
        overall\: 5 \\
        review\_title\: Five Stars \\
        review\_text\: Nice trays to get organized. We buy these for work, and they work quite nicely. No complaints here! \\
"..." \\

\\
\\ "..." \\

{\color{darkgreen}Sequence Item Position\: 6 }\\
Final Item\: True \\
Evaluate purchase intent of the following product \\
        item\_title\: Avery Two-Pocket Folders, Green, Pack of 25 (47977) \\
        item\_desc\: Fasteners for punched sheets, pockets for loose sheets--covers it all! Dual use makes it perfect for sales materials. Business card slot for convenient contact information. \\
        item\_price\: 17.219999313354492 \\
        overall\: 5 \\
        review\_title\: Five Stars \\
        review\_text\: Great presentation folders to have. I'll buy them again. \\
{\color{darkgreen} End of Sequence }\\
\\
\textbf{expected\_answer} \\
\{ \\
  "purchase reason"\: "Needed folders to store sales presentation materials, consisting of loose sheets and punched sheets.", \\
  "purchase reason explanation": "The customer needed a solution that could store sales presentation materials consisting of loose sheets and punched sheets.", \\
  "brief user summary"\: "........", \\
  "long user summary"\: "........", \\
  "user profile"\: "........" \\
\} \\
\\
----------------------------------------------------------------------- \\
Please (1) analyze the purchase reason of the last product purchased (2) provide a user summary, given the customer's product purchase and review history. Make sure the purchase reason does not include post-purchase product experiences.\\

Current ID:\\
$<$current\_ID$>$\\
\\
Customer Item/Review:\\
$<$current\_reviews$>$\\
\\
\\
\textbf{Answer:}\\

\\ \hline 
\textbf{Prompt for product endorsement:}\\
You are an exceptional customer engagement specialist at Amazon. Your role is to understand customer purchase motivations and experiences.\\
You will be provided with a sequence of product purchases made by a user and their corresponding reviews.\\
**Each sequence is ordered chronologically, from earliest to most recent purchase, marked by 'Sequence Item Position'.**\\
Your task is to focus on the **last product** in their purchase sequence (marked as 'Final Item: True') and generate an endorsement for the last product using the user's purchase history, review history and product description.\\
\\
Provide the following information in JSON format:\\
 **product endorsement**: Generate a sales pitch for the last product using the product description, user's purchase history and review history.\\

\textbf{Example JSON Output\:}\\
json\\
\{\\
  "product endorsement"\: "........"\\
\}\\
Important:\\
DO NOT quote any part of the user's reviews verbatim in your answers.\\
Focus on insights, not direct repetition.\\
Be specific and relate to the context of each customer's purchase and review history.\\
-----------------------------------------------------------------------\\
Current ID:\\
$<$current\_ID$>$ \\
\\
Customer Item/Review:\\
$<$current\_reviews$>$

\textbf{Answer}:

\end{longtable} 
\twocolumn

\onecolumn
\begin{longtable}{|p{0.9\textwidth}|} 
\caption{Examples of Generated Narratives}
\label{tab:data_example}
\\
\hline \\

\textbf{Generated Output 1} \\

  \textit{"purchase reason"}: "The customer wanted a stylish, relatively small tote bag with multiple pockets for organization."\\
  \textit{"purchase reason explanation"}: "The customer had previously purchased a dress and was looking for a stylish handbag that would match. They mentioned the bag was smaller than expected, but was still stylish and had multiple pockets. This suggests the customer is looking for a functional and fashionable handbag."\\
  \textit{"brief user summary"}: "This customer is a woman who enjoys shopping for both stylish and comfortable clothing. She is also interested in fashion accessories such as handbags."\\
  \textit{"long user summary"}: "This customer is a woman who has a taste for stylish and comfortable clothing. She has purchased sandals, dresses, and underwear, all items that suggest a preference for fashion and comfort. Her reviews also mention that she values quality, especially in clothing items like dresses. She is also interested in fashion accessories such as handbags. She seems to be an occasional buyer with a more practical and less frequent shopping approach. "\\
  \textit{"user profile"}: "Fashionable and practical woman"\\
  \textit{"product endorsement"}: "This Michael Kors Bedford Top Zip Pocket Tote Bag is a luxurious and functional choice, perfect for someone who appreciates quality and style. Based on your previous purchases of  comfortable clothing and stylish sandals, this bag complements your taste for both everyday practicality and elegant accents. The durable leather, spacious interior with multiple pockets, and adjustable handles make it ideal for carrying your essentials in a sophisticated way.  It's a great investment in a timeless piece that will elevate your everyday looks and be a reliable companion for years to come."\\

\hline \\
\textbf{Generated Output 2} \\

  \textit{"purchase reason"}: "The customer wanted a new casual dress."\\
  \textit{"purchase reason explanation"}: "The customer has a history of purchasing various dresses and has also mentioned in a review that they like a dress that is flattering on a “buxom” body.  This purchase is likely for a casual dress that they want to wear." \\
  \textit{"brief user summary"}: "This customer buys a wide variety of clothing and accessories for herself, her children, and her husband. They often look for items with comfortable fit and durability."\\
  \textit{"long user summary"}: "This customer buys a wide variety of clothing and accessories for herself, her husband, and children. She often seeks items that are comfortable and durable. She's interested in items that are well-made and have good value for money. For herself, she enjoys stylish and functional items, especially dresses, tops, and accessories. She also prioritizes practicality and comfort in clothing for her kids. Her reviews often highlight comfort, fit, and durability as important factors, particularly when it comes to clothing for children. Her purchase history reveals a preference for clothing that suits various occasions and personal styles. She likes to express her personal style through clothing and accessories and is not afraid to experiment with different looks."\\
  \textit{"user profile"}: "Fashionable, practical, family-oriented shopper"\\
   \textit{"product endorsement"}: "This casual A-line dress is perfect for a variety of occasions, from a casual day out to a more formal event. With its cap sleeves and V-neck, it is both stylish and comfortable. The dress is made from a lightweight and breathable fabric, making it perfect for warmer weather. And with its flattering A-line cut, it will look great on everyone.  You've previously purchased a variety of clothing items, ranging from formal dresses to everyday wear, and appreciated the quality of pieces that fit well. This dress is a great option for those who appreciate comfortable yet stylish clothing."\\

\hline \\
\textbf{Generated Output 3} \\

  \textit{"purchase reason"}: "The customer was looking for a short-sleeve fishing shirt that is breathable and provides UV protection."\\
  \textit{"purchase reason explanation"}: "The customer purchased a similar fishing shirt previously and rated it 5 stars for its fit, quality and features. The review mentions that the shirt is breathable, dries fast and provides UV protection."\\
  \textit{"brief user summary"}: "This customer is a man who is looking for comfortable and well-made work wear, specifically work boots and fishing shirts. He also values good quality and is looking for items that will last."\\
  \textit{"long user summary"}: "This customer is a man who seems to enjoy outdoor activities such as fishing, as he has bought multiple fishing shirts. The customer is also interested in finding comfortable and well-made workwear, having purchased work boots and a work shirt. He is willing to pay for quality and looks for items that will last. While he seems to be satisfied with the quality of the fishing shirts, he doesn't seem to appreciate the quality of a work shirt he purchased before."\\
  \textit{"user profile"}: "Outdoor enthusiast, workwear shopper"\\
    \textit{"product endorsement"}: "This Columbia Men's PFG Bahama II Short Sleeve Breathable Fishing Shirt is a great choice for anyone who wants a comfortable and durable shirt for fishing or other outdoor activities. You've already shown a preference for high-quality, comfortable clothing, like the KEEN Utility Men's Pittsburgh Steel Toe Work Boot, and this shirt is no different. The shirt is made of lightweight, durable nylon, dries quickly, and has mesh-lined cape vents at the back shoulder to keep you cool. It also has built-in UV protection, making it a great choice for sunny days on the water. Plus, you've already experienced the quality of Columbia clothing with your previous purchase, so you can be confident in the quality of this shirt."\\
\hline \\
\textbf{Generated Output 4} \\
  \textit{"purchase reason"}: "The customer needed a comfortable running shoe with a good fit."\\
  \textit{"purchase reason explanation"}: "The customer needed a new pair of running shoes, and had to replace the original insole due to discomfort on the sides. They found that an insole from their old Saucony shoes solved the problem, making the new shoes comfortable."\\
  \textit{"brief user summary"}: "The customer is a diverse shopper interested in men's clothing, travel accessories, and women's shoes and lingerie. They have purchased items for themselves and others." \\
  \textit{"long user summary"}: "This customer enjoys shopping for various items, including men's neckties and socks, travel accessories, women's lingerie, and women's running shoes. They have purchased items for both themselves and others, such as the socks for a friend. The customer has expressed a preference for comfortable and well-fitting items, as seen in their reviews for the shoes and lingerie. They are also drawn to humorous and unique items, as demonstrated by the purchase of the 'Nuclear Physics' necktie." \\
  \textit{"user profile"}: "Diverse, Practical, Comfort-Oriented" \\
  \textit{"product endorsement"}: "Looking for a running shoe that offers maximum comfort and energy return? The Saucony Women's Triumph ISO 3 Running Sneaker is designed to elevate your run with its increased Everun landing zone and dynamic ISOFIT system. This shoe provides a comfortable and supportive fit, adapting to your stride for a smooth and enjoyable running experience.  You've previously purchased items for yourself and others, focusing on both comfort and style, so you'll appreciate the Triumph ISO 3's blend of performance and everyday wearability. You've also enjoyed durable and comfortable items, and the Triumph ISO 3 is made with high-quality materials to ensure long-lasting performance.  Treat yourself to the ultimate running experience with the Saucony Women's Triumph ISO 3 Running Sneaker." \\
\\ \hline 
\end{longtable} 
\twocolumn

\onecolumn
\begin{longtable}{|p{0.9\textwidth}|} 
\caption{Example Auto-Rater prompt used for evaluation (Rating Veracity on User Summaries)}
\label{tab:rater_prompt}
\\
\hline 
You are a highly skilled customer behavior analyst AI. \\

You will be provided with user purchase history, including the item details, the review text. \\

Your task is to answer questions about the user's purchase behavior. \\

{\color{darkgreen}--- BEGIN USER PURCHASE HISTORY --- }\\
User Purchase: On YYYY-MM-DD the following was purchased \\
\\
{\color{darkgreen}--- BEGIN RESULTS --- }\\
* Name: … \\
  Category: … \\
  Description: … \\
  Price: … \\
{\color{darkgreen}--- END RESULTS ---} \\

User Purchase Review: … \\
    
… \\

{\color{darkgreen}--- END USER PURCHASE HISTORY --- }\\
\\
\textbf{Evaluation Criteria:} \\
Your responses will be evaluated based on: \\
* Accuracy: how well your judgment aligns with the true reason for purchase (if available) \\
* Evidence: the quality and relevance of the examples cited from the review history \\
* Reasoning: the clarity and logic of your analysis \\
* Confidence: how well your confidence level reflects the strength of the evidence \\
 \\

\textbf{Instructions:} \\
* Analyze the user's review history carefully to identify patterns in preferences, needs, or dislikes \\
* Compare the suggested reason and explanation against those patterns \\
* Consider both positive and negative evidence \\
* Be explicit in citing specific examples from the review history to support your assessment \\
* If there's insufficient evidence to make a strong judgment, state so clearly and explain why \\
* **Strictly penalize any purchase reasons that imply future knowledge of the user's experience with the product** with very low rating. \\
\\
\textbf{Additional Considerations:} \\
* Pay close attention to the language used in user reviews, especially adjectives, adverbs, and any mentions of specific features or occasions. Analyze the sentiment expressed towards these aspects. \\
* Evaluate each element of the provided purchase reason individually. If the reason mentions specific attributes (e.g., color, size, material), look for direct evidence of those preferences in the purchase history.\\
* Prioritize user statements and reviews over product descriptions when evaluating the purchase reason. Only use product descriptions as supporting evidence if they are directly corroborated by user statements. \\
* Analyze seemingly unrelated purchases for potential connections to the current purchase reason. For example, a past purchase of a gift might reveal information about the user's gifting preferences. \\
* If evidence for a specific aspect of the reason is limited or based solely on assumptions, express lower confidence in the overall judgment. \\
\\
\textbf{Question:} \\
Please evaluate the veracity of the provided user profile, user brief summary, and user long summary in light of the user's purchase history. \\

{\color{darkgreen}--- BEGIN USER PROFILE --- }\\
…\\
{\color{darkgreen}--- END USER PROFILE --- }\\
\\
{\color{darkgreen}--- BEGIN USER BRIEF SUMMARY --- }\\
…\\
{\color{darkgreen}--- END USER BRIEF SUMMARY --- }\\
\\
{\color{darkgreen}--- BEGIN USER LONG SUMMARY --- }\\
…\\
{\color{darkgreen}--- END USER LONG SUMMARY --- }\\
\\
\textbf{Output:} \\
A detailed assessment of the veracity of the provided user profile, user brief summary, and user long summary. Your assessment should include: \\

* Clear judgment on whether the summaries and profile are true, false, or uncertain, based on the evidence from the review history. \\
* Specific examples from the review history that support your judgment. \\
    * You may use the entirety of the user's purchase history to validate the claims made in the summaries. \\
    * Keep in mind that even a negative review on a specific attribute can indicate that the user cares about that aspect. \\
    * Avoid making broad generalizations or inferences that are not directly supported by the evidence. \\
    * Focus on strong, direct evidence that clearly connects to the specific claims in the summaries. \\
    * Analysis of any potential inconsistencies or contradictions between the summaries and the user's past  behavior. \\
* Consideration of the full range of evidence, including potentially contradictory information. \\
* Acknowledgment of complexities and uncertainties if present. \\
* Overall confidence level in your assessment (high, medium, low). \\

Remember, your end judgment must be one of: \\
\\
 - \textit{Likely Very False}  \\
 - \textit{Likely False} \\
 - \textit{Likely Somewhat False} \\
 - \textit{Uncertain} \\
 - \textit{Likely Somewhat True} \\
 - \textit{Likely True} \\
 - \textit{Likely Very True} \\
\\
Please think aloud about how would you answer this question and provide your reasoning prior to answering. \\

Answer: \\

\\ \hline 

\end{longtable} 
\twocolumn

\definecolor{darkgreen}{rgb}{0.0, 0.5, 0.0} 


\begin{table*}[ht]
\centering
\begin{tabular}{|p{5cm}|p{5cm}|p{4cm}|} 
\hline
\textbf{Reference} & \textbf{Prediction} & \textbf{Comments} \\
\hline
\rowcolor{green!15}
To replenish their supply of label maker tape.	& To replenish their supply of DYMO label maker tape, specifically red print on white tape & \multirow{2}{4cm}{Mostly inline with the reference.} \\
\cline{1-2}
\rowcolor{green!15}
To enhance the user's writing experience with high-quality pencils, complementing their existing collection of premium writing instruments. &	To obtain a high-quality pencil sharpener for personal use, possibly for colored pencils. & \\
\rowcolor{green!15}
\cline{1-2}
\rowcolor{green!15} The customer needed large index cards for note-taking and drawing, preferring a color that is easy on the eyes. &	The customer needed high-quality, heavy-weight index cards for note-taking and flashcards.
Replacement of a previously owned 3M Precise mouse pad.	Replacement of a mousepad. & \\
\cline{1-2}
\rowcolor{green!15} To organize schoolwork. &	To organize their children's school binders. & \\
\hline
\rowcolor{yellow!10} To further organize their desk	& To further organize their desk and potentially store business cards. & Model may be hallucinating. (store business cards) \\
\hline
\end{tabular}
\caption{Examples of generated purchase reasons compared to reference reasons.}
\label{tab:pr_traces}
\end{table*}

\begin{table*}[ht]
\centering
\begin{tabular}{|p{5cm}|p{5cm}|p{4cm}|} 
\hline
\textbf{Reference} & \textbf{Prediction} & \textbf{Comments} \\
\hline
\rowcolor{green!15} The customer's purchase history shows a focus on office and school supplies, particularly items used for organization and labeling. The purchase of a bulk pack of lanyards with ID holders suggests a need for efficient identification and organization of multiple individuals, aligning with a camp or similar event scenario. & The customer's purchase history shows a pattern of buying office and school supplies, including items for organization and labeling. The purchase of a bulk pack of lanyards and badge holders strongly suggests the need for these items for a large event, such as a conference or field trip. & \multirow{2}{4cm}{Reference is more detailed in capturing the richness of history, whereas this model is more succinct in its output, capturing only the essence of the history.} \\ 
\cline{1-2}
\rowcolor{green!15} The customer has a history of purchasing various fountain pens and related supplies, indicating an interest in this writing instrument. The purchase of disposable pens suggests a need for a less expensive option, possibly for practice or projects where higher-quality pens might be unsuitable. The product description mentions calligraphy, further supporting this assumption. &	The customer has a history of purchasing various pens and refills, showing an interest in writing instruments. The low price point of the Pilot Varsity pens, combined with the previous purchases of Jinhao and other brand fountain pens, suggests a desire to explore different pen types and potentially experiment with different brands before potentially investing in higher-quality options. & \\ 
\hline
\rowcolor{yellow!10} The customer's purchase history shows a preference for Lexmark ink cartridges. Their final purchase of a tri-pack containing two black and one color cartridge indicates a need for replacement ink, with a higher demand for black ink than color ink. & The customer has previously purchased Lexmark ink cartridges (both black and color) and given positive reviews. The final purchase is another Lexmark ink cartridge, this time with a three-pack of black, cyan, magenta, and yellow, indicating a need to replenish their supply of color ink for their Lexmark printer. &	Reference is more precise in parsing the nuance of tri-pack containing 2 Black and 1 Color cartridge. \\ 
\hline
\rowcolor{red!10}
The customer previously purchased a rubber band ball and gave it a 5-star rating, indicating satisfaction with the product type for organizing rubber bands. The positive review of the previous rubber band ball, coupled with the description of the current purchase as an item the customer reorders as needed, strongly suggests the purchase is for the same organizational purpose. &	The customer previously purchased a similar item ('Ball Ornament Peanuts') and described it as a 'good addition to my service bag'. The purchase of this second item, 'Just Peanuts', suggests a continued need for this type of organizational tool in their work or home office. & This model hallucinates 'Ornament Peanuts'. \\ 
\hline
\end{tabular}
\caption{Examples of generated purchase reason explanations compared to reference explanations.}
\label{tab:pre_traces}
\end{table*}


\begin{table*}[ht]
\centering
\begin{tabular}{|p{5cm}|p{5cm}|p{4cm}|} 
\hline
\textbf{Reference} & \textbf{Prediction} & \textbf{Comments} \\
\hline
\rowcolor{green!15} This customer is a discerning buyer who values both functionality and quality in their office supplies. Their purchase history shows a focus on high-quality writing instruments, as evidenced by their purchase of both a Rotring 800 and subsequently a Rotring 600 mechanical pencil. The purchase of a refillable eraser holder demonstrates a preference for durable and long-lasting items. Furthermore, their acquisition of index dividers and a binder suggests a need for effective organization. The purchase of a toner cartridge points towards the need to refill their printer. The overall shopping pattern suggests a preference for well-designed, reliable tools for their academic or professional work, possibly someone who values precision and efficiency.	
& This customer demonstrates a clear preference for high-quality office supplies, particularly writing instruments and organizational tools. Their purchases include a premium eraser, legal-size dividers, a binder (though they were dissatisfied with the quality), and toner cartridges. Their final purchase, a large, high-end mechanical pencil, shows a focus on comfort and durability, possibly for a work profession where they value writing instruments' longevity. The customer also seems to prioritize value for money, as evidenced by their toner cartridge purchase. While they are willing to spend more on items that they believe are of superior quality and long-term value, their purchase of the mechanical pencil suggests that they are also willing to pay more for an improvement in a specific product.	
& This model shows good capabilities matching the Gemini output in detail and descrption \\

\hline
\rowcolor{yellow!10}
The customer's purchase history reveals a preference for high-quality writing instruments suitable for crossword puzzles. They have purchased various pens, including gel pens, rollerball pens, and even a tactical pen, indicating a focus on writing experience and ink quality. Their need for durable office supplies is highlighted by the negative review of the thin, easily cracked clipboard; they explicitly mention the need for a sturdier model to withstand their intended use. The purchase of high-quality tape suggests a need for reliable adhesive supplies for organizational purposes. Their purchase of a large gaming mouse pad suggests the customer potentially engages in gaming as well.	
& This customer's purchase history reveals a focus on quality writing instruments and office supplies. They have purchased high-end pens (Pilot Precise V7), pen refills, and a high-performance mousepad, indicating a preference for smooth writing and a comfortable workspace. The purchases of tape and clipboards further suggest a need for practical and durable office tools. The customer appears to prioritize value and functionality, seeking products that meet their specific needs, even if they are not averse to more expensive or specialized items.	& This model sometimes misses nuances. For eg, this customer is a gamer and the model is not able to pick that attribute \\
\bottomrule
\end{tabular}
\caption{Examples of generated (long) user summaries compared to reference summaries.}
\label{tab:us_traces}
\end{table*}

\begin{table*}[ht]
\centering
\begin{tabular}{|p{5cm}|p{5cm}|p{4cm}|} 
\hline
\textbf{Reference} & \textbf{Prediction} & \textbf{Comments} \\
\hline
\rowcolor{green!15} This customer frequently purchases printer toner and drum units for Brother laser printers, seeking both genuine and compatible options for cost-effectiveness. Their purchase history reflects a desire for high-yield cartridges and reliable performance.	& 
This customer frequently prints and uses a Brother printer,. They are looking for cost-effective printing solutions and have tried both Brother and compatible toner cartridges.	
& \multirow{2}{4cm}{This model shows good capabilities matching the reference output in detail and description.} \\
\cline{1-2}
\rowcolor{green!15}
This customer frequently buys office supplies and organizational tools, with a preference for quality products even if they are not always the most expensive. Their recent purchase of erasable colored pencils also suggests an interest in artistic pursuits. &	This customer prioritizes office organization and efficiency, purchasing various storage solutions, desk organizers, and printer supplies. They are also interested in art supplies, particularly colored pencils. & \\
\cline{1-2}
\rowcolor{green!15} This customer designs board games and frequently uses Sharpie markers of various colors and point sizes. They also utilize Elmer's glue for crafting and assembling game components. & This customer enjoys arts and crafts, particularly creating booklets, and prefers Sharpie brand markers for their vibrant colors and fine point. & \\
\bottomrule
\end{tabular}
\caption{Examples of generated (brief) user summaries compared to reference summaries.}
\label{tab:brief_us_traces}
\end{table*}

\end{document}